\documentclass[letterpaper]{article} 
\usepackage{sty/aaai2026}  
\usepackage{times}  
\usepackage{helvet}  
\usepackage{courier}  
\usepackage[hyphens]{url}  
\usepackage{graphicx} 
\usepackage{natbib}  
\usepackage{caption} 

\usepackage[T1]{fontenc}
\usepackage{multirow}
\usepackage{ifthen}
\usepackage{algorithm}
\usepackage[noEnd=true,indLines=true,commentColor=black]{sty/algpseudocodex}
\usepackage{stmaryrd}
\usepackage{amsmath,amssymb,amsthm}
\usepackage{booktabs}
\usepackage{cleveref}
\usepackage{siunitx}
\usetikzlibrary{decorations.pathmorphing}
\usepackage{sty/custom}
\usepackage{IEEEtrantools}
\usepackage{newfloat}
\usepackage{listings}
\DeclareCaptionStyle{ruled}{labelfont=normalfont,labelsep=colon,strut=off} 
\floatstyle{ruled}
\newfloat{listing}{tb}{lst}{}
\floatname{listing}{Listing}
\title{Graph Attention-Guided Search for Dense Multi-Agent Pathfinding}
\author{Rishabh Jain$^{1\ast}$, Keisuke Okumura$^{1,2\ast}$, Michael Amir$^1$, Amanda Prorok$^1$}
\affiliations{
  $^1$University of Cambridge, UK
  \\
  $^2$National Institute of Advanced Industrial Science and Technology (AIST), Japan
  \\
  \texttt{\{rj412,ko393,ma2151,asp45\}@cst.cam.ac.uk}
}
\nocopyright

\begin{document}

\maketitle
\let\thefootnote\relax\footnotetext{$^\ast$These authors contributed equally to this work.}

\begin{abstract}
Finding near-optimal solutions for dense multi-agent pathfinding (MAPF) problems in real-time remains challenging even for state-of-the-art planners.
To this end, we develop a hybrid framework that integrates a learned heuristic derived from MAGAT, a neural MAPF policy with a graph attention scheme, into a leading search-based algorithm, LaCAM. 
While prior work has explored learning-guided search in MAPF, such methods have historically underperformed. 
In contrast, our approach, termed \lagat, outperforms both purely search-based and purely learning-based methods in dense scenarios.
This is achieved through an enhanced MAGAT architecture, a pre-train–then–fine-tune strategy on maps of interest, and a deadlock detection scheme to account for imperfect neural guidance.
Our results demonstrate that, when carefully designed, hybrid search offers a powerful solution for tightly coupled, challenging multi-agent coordination problems.
\end{abstract}


\section{Introduction}
\emph{Multi-agent pathfinding (MAPF)} is a fundamental problem in multi-robot coordination, where the goal is to compute collision-free paths that efficiently guide a team of agents to their respective destinations.
Since the early 2010s, driven by applications in warehouse automation, MAPF has been extensively studied in the AI planning community.
More recently, its well-defined and discrete problem structure has made MAPF an attractive benchmark for advanced machine learning techniques tailored to multi-agent settings~\cite{prorok2021holy,alkazzi2024comprehensive}.
These two streams form the main research thrusts in MAPF today:
\begin{itemize}
\item Developing real-time \emph{centralized} planners based on heuristic search, aiming for reliable algorithms that offer scalability with respect to the number of agents and (near-)optimal solution quality;
\item Developing robust \emph{decentralized} neural policies that operate under local observations, learn complex agent interactions, and generalize to unforeseen situations---while still offering competitive performance and greater scalability than centralized approaches.
\end{itemize}

{
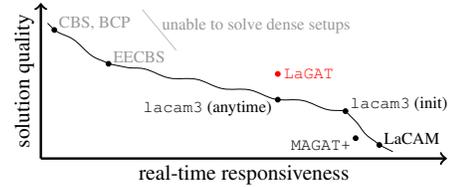
\begin{figure}[tp!]
\centering
\scalebox{0.9}{
\begin{tikzpicture}
  \scriptsize
  \draw[->, line width=1pt](0, 0) -- node[pos=0.5,above,rotate=90]{\footnotesize solution quality} (0, 2.3);
  \draw[->, line width=1pt](0, 0) -- node[pos=0.5,below]{\footnotesize real-time responsiveness} (6, 0);

  \coordinate (c0) at (5.2, 0.1);
  \coordinate (c1) at (5.0, 0.2);  
  \coordinate (c2) at (4.5, 0.7);  
  \coordinate (c3) at (1.0, 1.4);  
  \coordinate (c4) at (0.2, 1.9);
  \coordinate (c5) at (0.1, 2.0);

  \foreach \i/\j in {c0/c1, c1/c2, c2/c3, c3/c4, c4/c5} {
    \draw[decorate,decoration={snake,amplitude=0.3mm,segment length=8mm}]
    (\i) -- (\j);
  }

  \foreach \p/\x in {c1/{LaCAM}, c2/{\lacamthree (init)}, c3/{\gl{EECBS}}, c4/{\gl{CBS, BCP}}} {
    \filldraw (\p) circle (1pt);
    \node[anchor=west] at ($(\p) + (-0.02, 0.1)$) {\x};
  }

  \draw[lightgray]($(c4) + (1.3, 0.3)$) -- ++(0.5, -0.6)
  node[pos=0.4,anchor=west]{\gl{unable to solve dense setups}};

  \coordinate (c-lagat-tuned) at (3.5, 1.25);
  \filldraw[red] (c-lagat-tuned) circle (1pt);
  \node[anchor=west,text=red] at ($(c-lagat-tuned) + (0, 0)$) {{\lagat}};
  %
  \coordinate (c-magat) at (4.65, 0.3);
  \filldraw (c-magat) circle (1pt);
  \node[anchor=east] at ($(c-magat) + (0.02, -0.1)$) {\magatstar};
  %
  %
  \filldraw ($(c-lagat-tuned) + (0, -0.38)$) circle (1pt);
  \node[anchor=east] at ($(c-lagat-tuned) + (0.0, -0.5)$) {\lacamthree (anytime)};
\end{tikzpicture}
}
\caption{Pareto frontier consisting of state-of-the-art algorithms in dense MAPF instances.}
\label{fig:tradeoff}
\end{figure}
}

While both fields have made notable progress, learning-based approaches still lag significantly behind their search-based counterparts.
Across \emph{all} aspects, including scalability, leading search-based planners such as \lacamthree~\cite{okumura2024lacam3} consistently and significantly outperform existing learning paradigms~\cite{skrynnik2025pogema,andreychuk2025mapf}.
At the same time, as seen in various domains---most notably the game of Go~\cite{silver2016mastering,silver2017mastering}---the combination of search and learning has proven to be a powerful strategy, suggesting that a similar integration could lead to more capable MAPF methods.
Indeed, numerous efforts have been made to enhance classical MAPF search algorithms using neural components~\cite{yu2023accelerating,veerapaneni2024improving,yan2024neural,wang2025lns2+,alam2025optimizing}.
These approaches have succeeded in improving specific aspects of base planners; however, when considering factors such as real-time responsiveness, performance stability, scalability, and comparison with state-of-the-art planners, their impact remains limited---providing insight but not competitive performance.
One of these even concluded that \emph{``all learnt MAPF works require non-trivial data collection and complex models but perform worse than LaCAM''}~\cite{veerapaneni2024improving}.
Another benchmark paper similarly reported that \emph{``our results demonstrate that current learning-based methods fail to exhibit a clear advantage over simple rule-based heuristics''}~\cite{tan2025reevaluation}.

These insights raise practical questions in MAPF:
\emph{Can neural components actually provide benefits over state-of-the-art search-based planners, and if so, how?}
In this study, we demonstrate that, in densely populated MAPF instances, a hybrid approach combining search and learning can outperform purely search-based methods in terms of solution quality.
Our approach, named \lagat, involves a computational overhead due to neural network inference;
however, we find that a substantial improvement in solution quality justifies this overhead.
As illustrated in \cref{fig:tradeoff}, this work marks a new frontier---\emph{arguably for the first time by neural methods}---surpassing the existing Pareto frontier established by leading search-based algorithms.
Together with the recent evolution of the hybrid planner in lifelong MAPF~\cite{jiang2025deploying}---where tightly coupled coordination is not critical unlike dense one-shot MAPF setups---our findings provide a strong incentive to further pursue this direction.

\lagat embeds \magatstar, a lightweight neural heuristic, into the search-based MAPF algorithm \emph{LaCAM}~\cite{okumura2023lacam}.
\magatstar is our improved version of MAGAT~\cite{li2021message} (a seminal decentralized neural MAPF policy), attaining similar performance to leading neural methods while running faster and requiring less data to train.
We first pre‑train \magatstar by imitating \lacamthree trajectories, then fine‑tune it on the map of interest using a small amount of data.
Although the one-off adaptation might appear to constrain flexibility, it is a logical strategy in practice, given that major MAPF applications (e.g., warehouse automation and railway scheduling) assume static map layouts.
As neural policies alone are prone to inaccuracies and deadlocking, we cover for \magatstar's weaknesses using \emph{(i)}~\emph{PIBT} collision shielding~\cite{okumura2022priority} and \emph{(ii)}~a novel deadlock detection mechanism.
The resulting \lagat surpasses the real-time performance of \lacamthree in dense setups.
Our code is available in \url{https://github.com/proroklab/lagat}.

\section{Preliminaries}
Our proposed \lagat is built upon PIBT and LaCAM---two high-performing algorithms that currently lead research on real-time and scalable MAPF---together with MAGAT.
Starting from the problem formulation, this section outlines PIBT, LaCAM, and neural MAPF policies, including MAGAT, which serve as the foundation for the rest of this paper.

\subsection{Problem Definition and Notations}
We consider a widely used MAPF formulation~\cite{stern2019def}, defined by a team of agents $A = \{1, 2, \ldots, n\}$ a four-connected grid graph $G=(V, E)$, and a distinct start $s_i \in V$ and goal $g_i \in V$ for each agent $i \in A$.
At each timestep, each agent either stays in place or moves to an adjacent vertex.
Vertex and edge collisions are prohibited: two agents cannot occupy the same vertex simultaneously, and two agents cannot swap their occupied vertices within one timestep.
Then, for each agent $i \in A$, we aim to assign a collision-free path, denoted by $\pi_i = (v^0 = s_i, v^1, \ldots, v^T = g_i)$.
The solution quality is assessed by \emph{sum-of-costs} (SoC; aka. flowtime), which sums the travel time of each agent until it stops at the target location and no longer moves.

Herein, a \emph{configuration} $\Q \in V^n$ refers to the locations for all agents.
We use $\bot$ as an ``undefined'' sign.
The \dist function returns the shortest path length between two vertices, while $\neigh(v)$ represents adjacent vertices for $v \in V$.

\subsection{PIBT and LaCAM}
\emph{PIBT (priority inheritance with backtracking)}~\cite{okumura2022priority} is a computationally lightweight function to generate a \emph{transitionable} configuration $\Q'$ given another configuration \Q, where each agent $i$ can move from the current location $\Q[i]$ to the next $\Q'[i]$ within one action, without colliding.
It has gained notable popularity as fast configuration generation is central to MAPF~\cite{jiang2025deploying,yukhnevich2025enhancing}.
For each generation, PIBT needs a \emph{preference} for each agent $i$, a sorted list of candidate actions $v \in \neigh(\Q[i]) \cup \{ \Q[i] \}$.
This is arranged in an ascending order with $\dist(v, g_i)$, called \emph{cost-to-go}, which encourages agents to head towards their respective goals.

PIBT alone functions only as a greedy collision avoidance planner, often leading to deadlocks or livelocks.
\emph{LaCAM (lazy constraints addition search)}~\cite{okumura2023lacam} improves upon this by adding a systematic search over configurations, which derives a transitionable configuration sequence that connects the start and goal.
As enumerating all successor configurations requires exponential time for the number of agents, LaCAM uses PIBT to guide the search, creating only one successor for each search node expansion.
Other successors are gradually synthesized by adding \emph{constraints}, which prescribe the next location of each agent.
This lazy successor generation significantly reduces planning efforts, rendering LaCAM a reliable, real-time, and scalable MAPF solver~\cite{zhang2024planning,shankar2025lf}.

\Cref{algo:lacam} presents a simplified pseudocode of LaCAM;
disregard \cref{algo:lacam:dd} for now, which will be revisited in \cref{sec:search}.
As with most search methods, it proceeds the search by updating two data structures:
\emph{(i)}~an \open stack that stores search nodes, denoted as $\N$, and
\emph{(ii)}~an \explored table that stores already discovered configurations.
Each search node also maintains constraints, denoted as \C, that are used for constrained configuration generation with PIBT (\cref{algo:lacam:config-generator}).
Initially, the constraints include nothing (denoted $\C\init$), allowing PIBT to generate an arbitrary configuration.
They are gradually developed for each node invocation (\cref{algo:lacam:lowlevel-search}; abstracted for brevity), eventually specifying exactly one configuration without freedom.
Once the search hits the goal, LaCAM extracts a solution by backtracking ancestor nodes (\cref{algo:lacam:backtrack});
otherwise, \open eventually gets empty, signifying the unsolvability of the instance (\cref{algo:lacam:unsolvable}).

LaCAM is a \emph{complete} algorithm for MAPF, i.e., it finds a solution if the problem is solvable; otherwise, it reports the non-existence.
Subsequent work~\cite{okumura2023lacam2} shows that slight adaptations can make it an anytime version \emph{LaCAM$^\ast$} that eventually converges on an optimal solution.

{
\renewcommand{\S}{\m{\mathcal{S}}}
\begin{algorithm}[t!]
\caption{LaCAM for MAPF}
\label{algo:lacam}
\begin{algorithmic}[1]
\small
\State initialize \open, \explored
\State $\N\init \leftarrow \langle
\config: \S = (s_1, \ldots, s_n),~
\tree: \llbracket~\C\init~\rrbracket
\rangle$
\label{algo:lacam:init-node}
\State $\open.\push(\N\init)$;~~$\explored[\S] = \N\init$
\While{$\open \neq \emptyset$}
\State $\N \leftarrow \open.\funcname{top}()$
\IfSingle{$\N.\config = (g_1, \ldots, g_n)$}{\Return $\backtrack(\N)$}
\label{algo:lacam:backtrack}
\IfSingle{$\N.\tree = \emptyset$}{$\open.\pop()$;~\Continue}
\State $\C \leftarrow \N.\tree.\pop()$
\State $\funcname{update\_constraints}(\N, \C)$
\label{algo:lacam:lowlevel-search}
\Comment{constraints synthesis}
\State $\Q\new \leftarrow \funcname{configuration\_generator}(\N, \C)$
\label{algo:lacam:config-generator}
\Comment{call PIBT}
\IfSingle{$\Q\new = \bot$}{\Continue}
\IfSingle{$\explored[\Q\new] \neq \bot$}{\Continue}
\State $\N\new \leftarrow \langle
\config: \Q\new,~
\tree: \llbracket~\C\init~\rrbracket
\rangle$
\State $\open.\push(\N\new)$;\;$\explored[\Q\new] = \N\new$
\State \textcolor{blue}{\Call{DeadlockDetection}{$\N, \Q\new$}}
\Comment{for neural policies}
\label{algo:lacam:dd}
\EndWhile
\State \Return \nosolution
\label{algo:lacam:unsolvable}
\end{algorithmic}
\end{algorithm}
}

\subsection{Preference Construction}
Although LaCAM has advanced the frontier of scalable MAPF, its solution is known to be highly suboptimal~\cite{shen2023tracking}.
This primarily stems from the myopic nature of PIBT, which lacks the ability to reason about long-term coordination beyond one timestep.
While LaCAM’s constraints help prevent local miscoordination such as deadlocks, they do not promote (near-)optimal coordination.

For more capable MAPF planners, it is crucial to embed advanced coordination reasoning in PIBT, corresponding to devising the preference construction currently determined by cost-to-go.
This leads to a line of research, such as congestion mitigation~\cite{chen2024traffic,zhang2024guidance} and tiebreaking~\cite{okumura2025lightweight}.
Among these, a promising strategy is to leverage offline experience to construct the preferences, i.e., imitation learning from (near\nobreakdash-)optimal MAPF solutions that were collected in a computationally intensive manner~\cite{veerapaneni2024improving,veerapaneni2025work,jiang2025deploying}.
This technique is also called \emph{collision shielding (CS-PIBT)} because it uses PIBT to protect the neural policy execution from inter-agent collisions.

This study also falls in the last category.
Within this scheme, prior work~\cite{veerapaneni2024improving} concluded that neural-enhanced LaCAM has difficulty outperforming purely heuristic counterparts due to imperfections in learned models and slower inference speeds.
In this work, however, we overturn this claim through careful architectural design.

\subsection{Neural Decentralized MAPF Policies}
Alongside the development of search-based MAPF methods, the recent machine learning revolution fueled by neural networks has also advanced learning-based approaches to MAPF~\cite{alkazzi2024comprehensive}.
In particular, motivated by scalable and adaptive deployments, research on decentralized MAPF policies has gained notable momentum.
Typically, these approaches aim to design a neural policy that interprets surrounding information---represented as a field of view (FOV)---and that selects the next action within a grid-world context, optionally assuming inter-agent communication.
Such policies are commonly trained using reinforcement learning~\cite{sartoretti2019primal} or imitation learning~\cite{andreychuk2025mapf}.
Graph neural networks (GNNs) are sometimes adopted as the backbone, as they naturally capture agent interactions through their message-passing structure.
Our study builds on a representative GNN architecture for MAPF, known as \emph{MAGAT (message-aware graph attention network)}~\cite{li2021message}.

\subsection{MAGAT}
A GNN-based MAPF planner receives a local observation for each agent $i$, encoded as a node feature $\mathbf{x}_i$ (optionally processed by a CNN), and operates over a communication graph \G, constructed based on spatial proximity within a radius $R\comm \in \mathbb{N}_{>0}$.
Each agent $i$ transmits a latent message $\mathbf{m}_i$ over \G, receives messages $\mathbf{m}_j$ from neighboring agents $j$, and computes its action based on the aggregated information.
This process is represented by a single GNN layer.
Finally, a multi-layer perceptron (MLP) decodes the embedding into an action distribution.

While earlier GNN-based policies, e.g.,~\cite{li2020graph}, aggregate incoming messages uniformly, MAGAT introduces a message-dependent attention mechanism that enables each agent $i$ to weigh the importance of messages from its neighbors $j$.
This prioritization improves generalization to larger grids and higher agent densities.

{
  \begin{figure*}[th!]
    \centering
    \includegraphics[width=1\linewidth,clip,trim={0 6.1cm 0 6.0cm}]{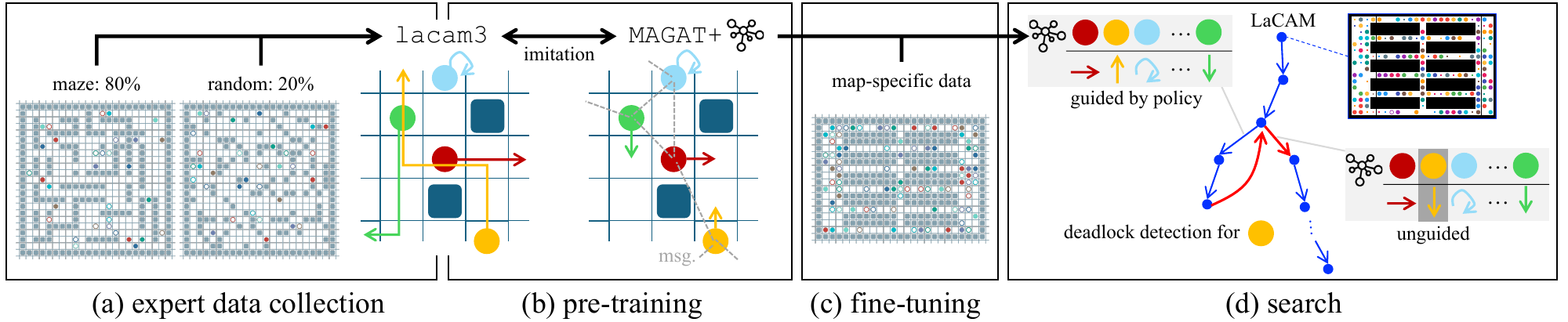}
    \caption{
      Overview of \lagat.
      We utilize 21K instances for pre-training (a, b) and 1K instances for fine-tuning (c).
      The resulting policy \magatstar is embedded into LaCAM search over configurations (d).
      When a deadlock is detected for a specific agent, the search performs a rollback, and the neural guidance is selectively overridden by the default, non-learning PIBT mechanism.
    }
    \label{fig:arch}
  \end{figure*}
}

\section{LaGAT}
The proposed \lagat is a \emph{map-specific}, complete, search-based MAPF solver that leverages neural policies as heuristics.
It integrates an enhanced version of LaCAM with an improved MAGAT policy, termed \magatstar, which injects long-horizon coordination into PIBT preferences through learned interaction reasoning.
We use MAGAT as our backbone because of its ability to capture complex agent interactions, which is paramount in dense setups.
As illustrated in \cref{fig:arch}, \lagat consists of four stages:
\emph{(i)}~collecting expert demonstration data using \texttt{lacam3},
\emph{(ii)}~pre-training \magatstar,
\emph{(iii)}~fine-tuning \magatstar to the target environment, and
\emph{(iv)}~executing LaCAM search guided by \magatstar, augmented with a newly developed deadlock detection mechanism.
In what follows, we describe the neural policy preparation and its integration into LaCAM in detail.

\subsection{Neural Policy Architecture (\magatstar)}

\paragraph{Local Observation.}
For each configuration $\Q$, each agent $i$ receives an observation tensor $o_i$ centered around its current position $\Q[i]$, with shape $4$$\times$$(2R\obs$$+$$1)$$\times$$(2R\obs$$+$$1)$, where $R\obs \in \mathbb{N}_{>0}$ determines the FOV size.
The first three channels are binary matrices indicating:
\emph{(i)}~the presence of obstacles,
\emph{(ii)}~the presence of other agents, and
\emph{(iii)}~a projection of the goal direction.
This design follows the observation architecture commonly used in neural MAPF policies~\cite{sartoretti2019primal,alkazzi2024comprehensive}.
The fourth channel, an addition to MAGAT, encodes cost-to-go values:
for each location $v \in V$ within the FOV, the value is computed as $(\dist(v, g_i) - \dist(\Q[i], g_i)) / (2R\obs)$.

\paragraph{Edge Feature.}
In the communication graph $\mathcal{G}$, each edge $(i, j)$ is associated with a three-dimensional feature vector $\omega_{ij}$, consisting of the relative position (in $x$ and $y$) and the Manhattan distance between the agents.
While the original MAGAT does not incorporate edge features, we empirically found that including them improves the model capability.

\paragraph{Model Architecture.}
\magatstar is an imitation learning policy $\pi_i(v \mid o_i, \G)$ that outputs an action distribution over candidate target vertices~$v$ for agent~$i$, given its observation~$o_i$ and the communication graph \G.
The model follows a similar design to the original MAGAT, consisting of
\emph{(i)}~a CNN encoder that converts $o_i$ into the node feature $\mathbf{x}_i$,
\emph{(ii)}~GNN layers, and
\emph{(iii)}~an MLP decoder.
Unlike the original MAGAT, which uses a single graph attention layer, \magatstar employs three stacked layers, to enable wider communication and greater cooperation via more message-passing interactions.
Another key modification is that these layers are extended to incorporate edge features, enabling richer interaction modeling.
Concretely, let $\mathbf{x}^{(l)}_i$ denote the node feature of agent~$i$ at layer~$l$, and $\omega_{ji}$ the edge feature from agent~$j$ to $i$.
The layer update proceeds with:

\begin{IEEEeqnarray}{rCl}
  \mathbf{x}^{(l+1)}_i & = & \sigma\left(\mathbf{W}_{\text{R}}^{(l)} \, \mathbf{x}^{(l)}_i + \mathbf{m}^{(l)}_{i} \right) \\
  \mathbf{m}^{(l)}_{i} & = & \sum_{j \in \mathcal{N}_i} \alpha^{(l)}_{ij} \, \left( \mathbf{W}_n^{(l)} \, \mathbf{x}^{(l)}_j + \mathbf{W}_e^{(l)} \, \mathbf{w}_{ji} \right) \\
  \alpha^{(l)}_{ij} & = & \frac{\exp\!\left(\mathrm{LeakyReLU}\left( a^{(l)}_{ij} \right)\right)}{\sum_{k \in \mathcal{N}_i} \exp\!\left(\mathrm{LeakyReLU}\left( a^{(l)}_{ik} \right)\right)} \\
  a^{(l)}_{ij} & = & \left({\mathbf{x}^{(l)}_i}\right)^{\top} \left( \Theta^{(l)}_n \, \mathbf{x}^{(l)}_j + \Theta^{(l)}_e \, \mathbf{w}_{ji} \right)
\end{IEEEeqnarray}
\noindent
Here, ${\mathbf{w}_{ji} = \phi(\omega_{ji})}$ is the processed edge feature via an MLP~$\phi$,
$\alpha_{ij}$ is the normalized attention weights,
$\mathbf{W}_{\{\text{R},\text{n},\text{e}\}}^{(l)}$ and $\Theta^{(l)}_{\{n,e\}}$ are the learnable weights,
and $\sigma$ is a non-linearity.
These enhancements allow GNN to capture more nuanced relational structures, improving policy performance in dense, complicated MAPF scenarios.

\subsection{Training Pipeline}

\paragraph{MAPF Instance Preparation.}
We use the POGEMA toolkit~\cite{skrynnik2025pogema}, also employed in the development of MAPF-GPT~\cite{andreychuk2025mapf}, a state-of-the-art imitation learning policy for MAPF.
Following MAPF-GPT's training protocol, we generate a total of 21K instances.
20\% feature randomly placed obstacles, while the remaining 80\% are maze-like environments (see~\cref{fig:arch}A).
Map sizes range from $17 \times 17$ to $21 \times 21$, and each instance includes 16, 24, or 32 agents.
For reference, MAPF-GPT uses 3.75M instances for training.

\paragraph{Collecting Expert Trajectories.}
To generate demonstration trajectories for training, we employ \lacamthree~\cite{okumura2024lacam3} instead of the conflict-based search variant~\cite{Barer-ECBS} used in MAGAT.
\lacamthree is an anytime planner that enables large-scale trajectory collection with low computational cost.
We adopt a staged timeout strategy with time limits of $[1, 5, 15, 60]\SI{}{\second}$.
If no solution is found within a given limit, the planner is re-run with the next longer timeout.
As the training set does not include challenging situations, most instances were solved within \SI{1}{\second}.

\paragraph{Pre-Training.}
\magatstar, with $R\comm=7$ and $R\obs=5$, is pre-trained on the collected expert trajectories using cross-entropy loss.
We train for 200 epochs, requiring about 100 hours on an NVIDIA L40S GPU, using the AdamW optimizer~\citep{Loshchilov-AdamW}.
Additional training details are provided in the appendix.
To mitigate the distributional shift common in imitation learning, we further apply on-demand dataset aggregation~\cite{ross2011reduction}, following~\citet{li2020graph}.

\paragraph{Map-wise Fine-Tuning.}
Given a target map $G$ (e.g., a warehouse map in \cref{fig:arch}), we generate additional $1000$ instances on $G$ and corresponding \lacamthree's trajectories, this time with higher agent densities: $32$, $48$, $64$, and $80$ agents.
These instances are used to fine-tune the pre-trained model, specializing it for $G$.
The fine-tuning process runs for 52 epochs and typically completes within 4–8 hours.

\subsection{Search with Imperfect Neural Policies}
\label{sec:search}

Neural MAPF policies alone, including \magatstar, are insufficient for solving MAPF, as their outputs are not guaranteed to be collision-free and therefore require additional safeguard mechanisms.
Moreover, such policies lack theoretical guarantees—such as completeness—which are often critical in real-world MAPF applications.
Integrating a non-learning search component is thus both practical and effective in addressing these limitations.

To this end, we adopt the strategy proposed by \citet{veerapaneni2024improving}, which uses PIBT to enforce safety and performs LaCAM on top.
In this setup, agent preferences in PIBT are constructed by sorting candidate vertices in descending order of the probabilities predicted by \magatstar.
Unlike the na\"{i}ve execution of neural policies that sample actions \emph{probabilistically}, this framework uses the model prediction \emph{deterministically} during the search.

Although our work builds on this concept, empirical results suggest that the vanilla neuro-enhanced approach does not fully realize the model’s potential to surpass existing MAPF solvers.
Several factors contribute to this limitation:
\emph{(i)}~Large models, such as those used in language modeling or MAPF-GPT,%
\footnote{
The smallest MAPF-GPT models require \SI{50}{\milli\second} for one timstep for 128 agents while \magatstar runs in $\leq$\SI{5}{\milli\second}.
}
are unsuitable for quick search due to their slow inference speed;
\emph{(ii)}~As a result, compact neural networks must be used.
  However, these lack sufficient representational power to mimic expert behavior accurately;
\emph{(iii)}~Inaccurate predictions lead to deadlocks or livelocks, especially with deterministic sampling employed, which significantly degrades LaCAM performance;
\emph{(iv)}~While probabilistic sampling improves robustness, it tends to execute suboptimal actions, reducing overall solution quality and limiting the benefits of hybridization.

In summary, \emph{we need to use neural policies deterministically, but they can misbehave occasionally}.
This motivates a systematic override mechanism that replaces model outputs with the default PIBT behavior when misbehavior is detected, which we refer to as \emph{deadlock detection}.

{
\begin{algorithm}[tp!]
  \small
\caption{\textsc{DeadlockDetection}}
\label{algo:dd}
\begin{algorithmic}[1]
\Input{search node \N, generated configuration $\Q\new$}
\Params{depth of deadlock detection $d \in \mathbb{N}_{\geq 0}$}
  \State $\N\ans \leftarrow \funcname{parent}(\N)$ \Comment{$\Q\ans \defeq \N\ans.\config$}
\For{$1, 2, \ldots, d$}~\textbf{until}~$\N\ans = \bot$
\For{$i \in A~\text{s.t.}~\Q\new[i] \neq g_i \land \Q\new[i] = \Q\ans[i]$}
\label{algo:dd:cond1}
\If{$\surrounding(\Q\new, i) = \surrounding(\Q\ans, i)$}
\label{algo:dd:cond2}
\State $\N\ans.\unguided.\funcname{insert}(i)$
\EndIf
\EndFor
\If{$\N\ans.\unguided$ has updated}
\State $\N\ans.\tree \leftarrow \llbracket~\C\init~\rrbracket$ \Comment{reset constraints}
\label{algo:dd:reset}
\State $\open.\push(\N\ans)$ \Comment{node reinsert}
\label{algo:dd:reinsert}
\EndIf
\State $\N\ans \leftarrow \funcname{parent}(\N\ans)$
\EndFor
\end{algorithmic}
\end{algorithm}
}

\subsection{Deadlock Detection}
A common failure of neural policies under deterministic sampling arises when an agent repeatedly executes the same action sequence without making progress toward its goal.
For example, we often observe oscillatory behavior between two positions, resulting in a livelock-like state.

LaCAM allows us to systematically detect such failure cases by backtracing recent agent-wise state histories.
Given a hyperparameter $d \in \mathbb{N}_{\geq 0}$ (typically less than 3), we check whether an agent has revisited the same state within the past $d$ timesteps.
If repetition is detected, the agent is deemed to be in deadlock or livelock, and its neural output is overridden with the default, non-learned preference.

\Cref{algo:dd} crystallizes this idea, embedded at \cref{algo:lacam:dd} in \cref{algo:lacam}.
For each search node in LaCAM, we maintain a set of \emph{unguided} agents by neural policies, initialized with $\emptyset$.
When an agent $i$ is detected to be in deadlock at a given configuration, it is added to the unguided set, thereby forcing it to follow the default preference.
The detection is performed at \cref{algo:dd:cond1,algo:dd:cond2}, by comparing its next location $\Q\new[i] \in V$ with the corresponding location in the ancestor state $\Q\ans[i]$, based on the following three conditions:
\emph{(i)}~the agent has not yet reached its goal,
\emph{(ii)}~the location remains unchanged, and
\emph{(iii)}~the vicinity remains the same, defined as:
$\surrounding(\Q, i) = \{ (v, j) \mid v \in \neigh(\Q[i]), \Q[j] = v \}$.
Once the unguided set is updated, \cref{algo:dd} reinserts the corresponding node into \open, allowing the next search iteration to explore it (\cref{algo:dd:reinsert}).
In addition, we discard the constraints explored so far (\cref{algo:dd:reset}) to eliminate unnecessary suboptimal behavior forced by the constraints.
{
  \newcommand{\tri}[1]{\textcolor[HTML]{#1}{\tiny\m{\blacksquare}}}
  \begin{figure}[tp!]
    \centering
    \scriptsize
    \begin{tikzpicture}
      \node[anchor=south west] at (0, 0) {\includegraphics[width=0.9\linewidth]
        {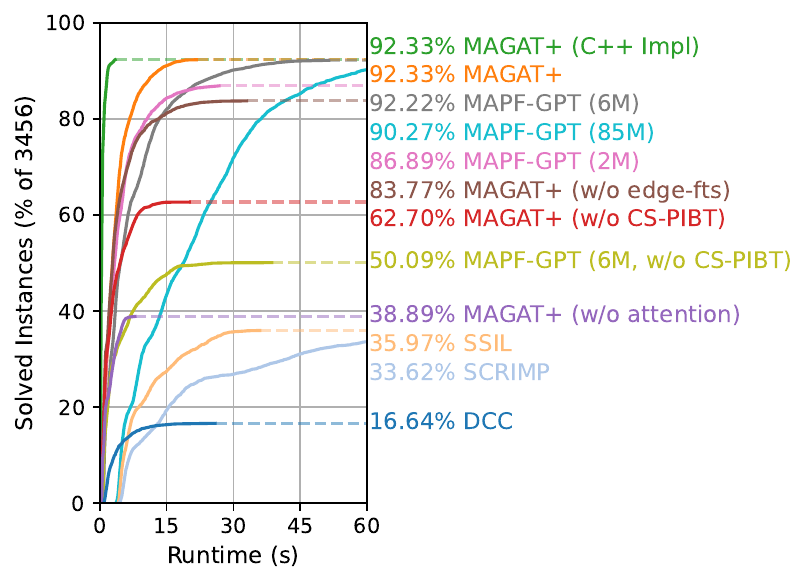}};
      \node[anchor=south west] at (5.14, 0.17) {
        \renewcommand{\arraystretch}{0.7}
        \setlength{\tabcolsep}{0.1mm}
        \begin{tabular}{rll}
          & Policy & \#Params \\
          \midrule
          \tri{ff7f0e}\tri{2ca02c} & \magatstar & 760K \\
          \tri{e377c2}\tri{7f7f7f}\tri{bcbd22} & MAPF-GPT & \{2,6,85\}M \\
          \tri{d62728}\tri{9467bd}\tri{8c564b} & Ab. \magatstar & 760K \\
          \tri{ffbb78} & SSIL & 160K \\
          \tri{aec7e8} & SCRIMP & 8M \\
          \tri{1f77b4} & DCC & 2M \\
        \end{tabular}
      };
    \end{tikzpicture}
    \caption{
      Evaluation on neural policies by success rate and runtime.
      Model sizes are listed in the bottom right corner.
    }
    \label{fig:success-by-runtime}
  \end{figure}
}


{
\newcommand{\fheight}{0.16\linewidth}
\newcommand{\fwidth}{0.16\linewidth}

\newcommand{\entry}[7]{
  \begin{scope}[xshift=#1*0.16\linewidth]
  \node[anchor=north west] at (0, 0)
       {\includegraphics[width=\fwidth,height=\fheight]{fig/raw/main_eval/#3_sum_of_costs}};
  \node[anchor=north west,fill=white,inner sep=0pt] at (0.60 + #6, -0.23)
       {\includegraphics[width=#7\linewidth]{fig/raw/main_eval/#3}};
  \node[anchor=center] at (1.65, -0.05) {\scriptsize\mapname{#2}};
  \node[anchor=west] at (0.5 + #6, -1.45) {\tiny #4$\times$#5};
  \node[anchor=north west] at (0, -2.8)
       {\includegraphics[width=\fwidth,height=\fheight]{fig/raw/main_eval/#3_per_map_runtime}};
  \end{scope}
}
\begin{figure*}[th!]
  \centering
  \begin{tikzpicture}
    \entry{0}{Dense Warehouse}{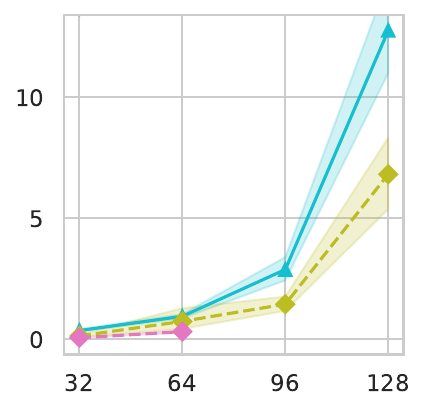}{23}{16}{0}{0.08}
    \entry{1}{Dense Room}{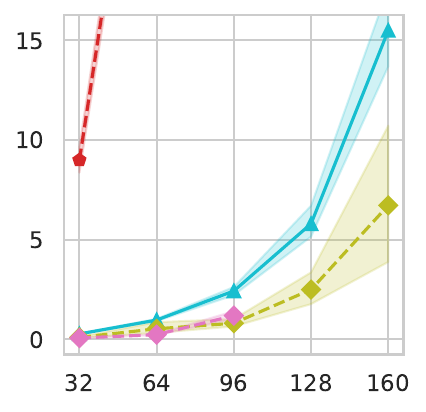}{23}{23}{-0.0}{0.06}
    \entry{2}{Dense Maze}{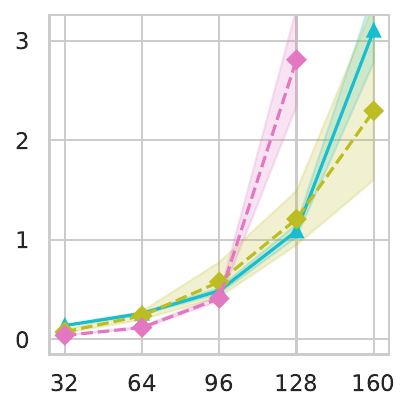}{21}{21}{-0.1}{0.06}
    \entry{3}{Empty Room}{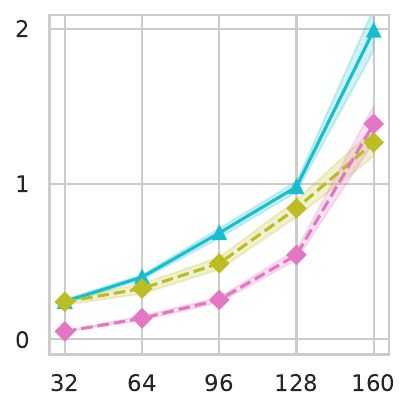}{23}{23}{0.05}{0.06}
    \entry{4}{ost003d}{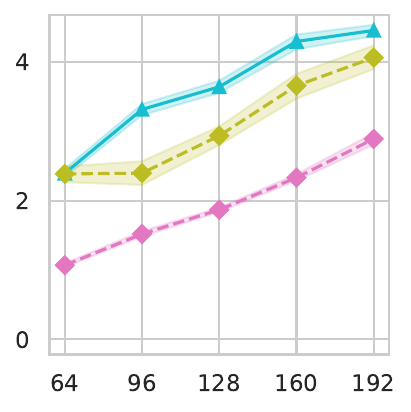}{194}{194}{0.05}{0.06}
    \entry{5}{Empty Map}{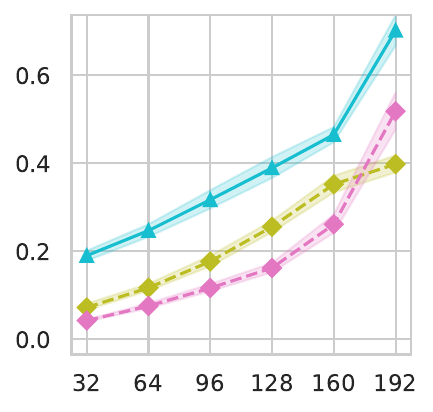}{20}{20}{0.1}{0.06}
    %
    \node[] at (0.1\linewidth, -6.0) {\small \#agents};
    \node[rotate=90] at (0, -1.5) {\small SoC / LB};
    \node[rotate=90] at (0, -4.3) {\small Runtime (\SI{}{\second})};
    \node[anchor=east] at (0.96\linewidth, -6.0)
         {\includegraphics[width=0.75\linewidth,clip,trim={0.4cm 0.3cm 0.4cm 0.3cm}]{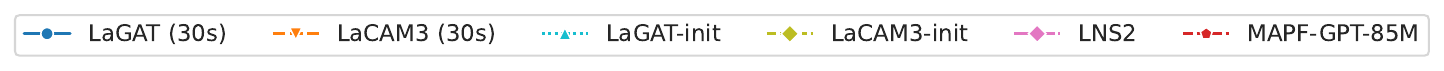}};
  \end{tikzpicture}
  \caption{
    Map-wise evaluation.
    Runtime and sum-of-costs scores are average within solved instances for each method.
    Transparent regions represent 95\% confidence intervals.
    SoC is normalized by its trivial lower bound $\sum_{i\in A}\dist(s_i, g_i)$.
    \mapfgpt is significantly slower than the other methods and falls outside the plotted runtime range.
    For reference, we also display the results of another leading search-based solver, LNS2~\cite{li2022mapf}.
    With a timeout of \SI{30}{\second}, LNS2 fails in dense setups (e.g., 0\% in \mapname{Dense
Warehouse} with 128 agents), so we only plot LNS2 data points with a success rate higher than 80\%.
  }
  \label{fig:main-eval}
\end{figure*}
}




\begin{theorem}
  LaCAM with \cref{algo:dd} is complete.
  \label{thrm:dd}
\end{theorem}
\noindent
The proof is available in the appendix.
The eventually optimal variant, LaCAM$^\ast$, remains readily attainable too.

\section{Evaluation}
We contrast \lagat with various state-of-the-art planners, comprising search-based solvers and neural policies based on either imitation or reinforcement learning:
\begin{itemize}
\item \lacamthree~\cite{okumura2024lacam3}, one of the most efficient, scalable and reliable search-based MAPF solvers available today.
\item \mapfgpt~\cite{andreychuk2025mapf}, the most capable imitation-learning policy to date, using GPT-like architecture.
  The authors provided three model sizes: 2M, 6M, and 85M.
  Generally, larger models are more capable.
\item \solver{SSIL}~\cite{veerapaneni2025work}, a follow-up of~\cite{veerapaneni2024improving} that uses GNN-based imitation learning.
\item \solver{SCRIMP}~\cite{Wang-SCRIMP} and \solver{DCC}~\cite{Ma-DCC}, reinforcement learning-based representatives.
\item \magatstar: our enhanced MAGAT model, without map-specific fine-tuning or search integration.
\end{itemize}
All implementations are obtained from the authors’ public repositories or the POGEMA benchmark suite~\cite{skrynnik2025pogema}.
\magatstar is developed and trained in Python, and translated into C++ for faster inference.
\lagat is also coded in C++.
Unless otherwise noted, all imitation learning methods (\mapfgpt, \solver{SSIL}, and \magatstar) are equipped with advanced collision shielding (CS-PIBT), as recommended in~\cite{veerapaneni2025work}.

\subsection{Neural Policy Performance}
\label{sec:result-policy}
We begin by evaluating the standalone performance of neural policies to justify the use of \magatstar as the learned component in \lagat.
\Cref{fig:success-by-runtime} shows a cactus plot illustrating the number of instances solved within varying time budgets, up to \SI{60}{\second}, across different map types and agent densities.
Specifically, we evaluate $3,456$ instances generated using POGEMA~\cite{skrynnik2025pogema}, covering six map types: \mapname{Warehouse}, \mapname{Room}, \mapname{Maze}, and their denser variants.
Further details about these maps are available in the appendix.
For each map, we consider four agent densities, ranging from $32$ to $160$ agents.
All experiments are conducted on a computing cluster equipped with Intel Xeon Platinum 8452Y CPUs and NVIDIA L40S GPUs for fast neural inference.
We also include several ablated versions of \magatstar to see their functionality.

From \cref{fig:success-by-runtime}, we first observe that imitation learning-based policies consistently outperform those trained via reinforcement learning.
The incorporation of CS-PIBT further boosts their effectiveness, aligning with prior findings~\cite{andreychuk2025mapf,veerapaneni2025work}.
Notably, \magatstar achieves success rates comparable to \mapfgpt, despite having a significantly smaller model size and being trained on a smaller dataset.
This strong performance is attributed to its carefully designed graph attention architecture.
Ablated versions confirm that each architectural component contributes meaningfully to the model’s overall capability.
Furthermore, \magatstar excels in inference speed;
particularly in our optimized C++ version, which significantly outperforms other models.
This efficiency is a critical feature for integrating neural policies into search-based MAPF solvers.
Armed with these insights, we next challenge \lacamthree.

\subsection{\lagat vs. \lacamthree}
\label{sec:result-main}
As \lacamthree is an anytime algorithm that progressively improves solution quality over time, we evaluate the planners’ real-time performance based on two criteria:
\emph{(i)}~quality of the initial feasible solution, and
\emph{(ii)}~final solution quality at the deadline, set to \SI{30}{\second}.
Since \lagat alone does not support anytime refinement, we incorporate a widely used post-processing method, large neighborhood search (LNS)~\cite{li2021anytime,okumura2021iterative}, which iteratively selects a subset of agents and refines their paths.
We adopt the same LNS implementation used in \lacamthree and attach it to \lagat to ensure a fair comparison.

Our evaluation includes both constrained challenging environments (e.g., \mapname{Dense Warehouse/Room}) and sparse, relatively easy ones (e.g., \mapname{ost003d}).
For each map and each density, 32 instances are prepared.
The experiment is conducted on a cluster with Intel Xeon Gold 6248R processors and NVIDIA GeForce RTX 2080 Ti GPUs.

\Cref{fig:main-eval} compares \lagat with leading search-based and learning-based solvers, namely \lacamthree and \mapfgpt.
All plotted results have a nearly 100\% success rate.
In dense environments (first three columns), although \lagat incurs non-negligible runtime for deriving initial solutions, it consistently produces significantly higher-quality initial solutions than not only \mapfgpt, but also \lacamthree.
As a result, it achieves markedly better final performance by the time limit---\emph{completely outperforming \lacamthree}.
\Cref{fig:scatter} demonstrates this conclusion.
The appendix contains success rates, and empirical results for additional maps.

We note that this advantage does not extend universally.
On larger, sparser maps such as \mapname{ost003d}, \lacamthree reliably generates near-optimal solutions, while \lagat tends to yield slightly suboptimal initial solutions.
Nonetheless, thanks to the efficiency of LNS in sparse settings, \lagat ultimately achieves comparable final performance.
Interestingly, we observe that \lagat can even slightly outperform in most cases;
we presume that \lagat solutions exhibit structural regularities that LNS can exploit more effectively.

\subsection{Analyzing Roles of Learning and Search}
\label{sec:result-ablation}
To better understand which technical components are responsible for the strong performance of initial solutions in the dense setups, the following ablated versions of \lagat are tested on the same experimental setup as in \cref{sec:result-main}:
\emph{(i)~w/o map-specific fine-tuning}: uses the pretrained policy without additional adaptation to the target map;
\emph{(ii)~w/o pre-training}: uses \magatstar trained only on data from the target map, without general pre-training;
\emph{(iii)~w/o neural guidance}: reverts to the original LaCAM;
\emph{(iv)~w/o deadlock detection}: disables \cref{algo:dd};
\emph{(v)~w/o search}: i.e., \magatstar with CS-PIBT.

\Cref{fig:ablation-bar-graph}, evaluated on the densest scenarios for the first three maps in \cref{fig:main-eval}, demonstrates that all components are essential for simultaneously achieving high success rates and low solution costs, underscoring the effectiveness of \lagat's design.
In particular, both pre-training and fine-tuning are critical for improving solvability, while neural guidance and search mechanisms, especially deadlock detection, contribute significantly to reducing solution cost.

{
  \newcommand{\hwidth}{0.45\linewidth}
  \begin{figure}[tp!]
    \centering
    \includegraphics[width=\hwidth]{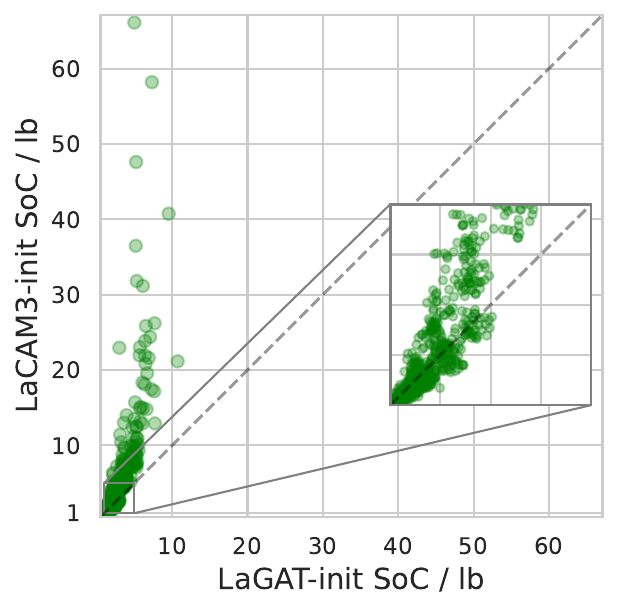}
    \includegraphics[width=\hwidth]{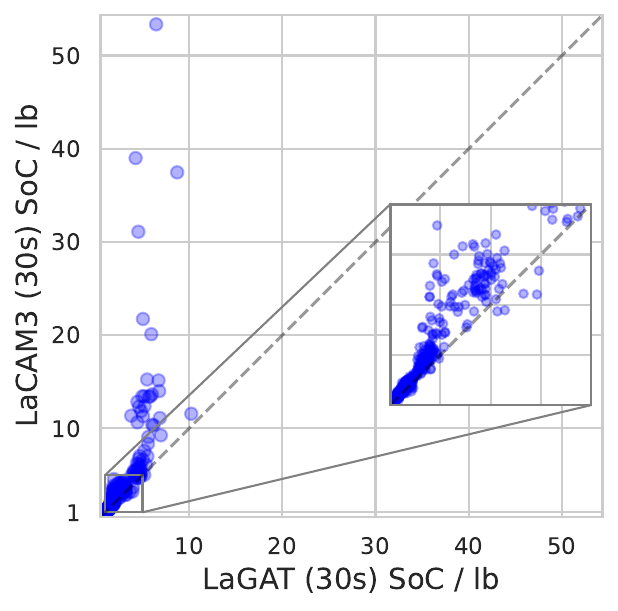}
    \caption{
    Instance-wise sum-of-costs comparison between \lacamthree and \lagat across all instances included in \cref{fig:main-eval}.
    Points in the upper-left triangular region indicate where \lagat outperforms \lacamthree.
    }
    \label{fig:scatter}
  \end{figure}
}

{
  \begin{figure}[tp!]
    \centering
    \includegraphics[width=1.0\linewidth]{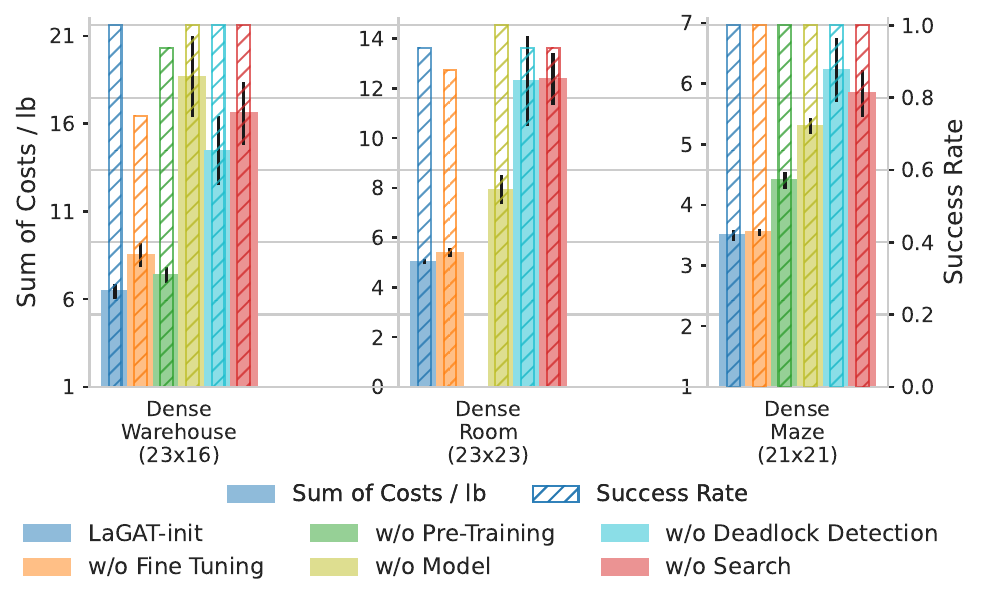}
    \caption{
      Ablation study on the densest scenarios in \cref{fig:main-eval}.
    }
    \label{fig:ablation-bar-graph}
  \end{figure}
}

\section{Discussion}
This study demonstrates that combining search with neural policies is effective in dense, highly congested MAPF scenarios, where purely search-based methods alone struggle to achieve strong results.

\paragraph{Why does LaGAT succeed?}
While prior work has attempted to embed neural policies into configuration-based planners---most notably \citet{veerapaneni2024improving}, who integrated MAGAT into LaCAM---the reported performance gains did not justify the runtime cost.
In contrast, our results demonstrate that, neural guidance can yield clear advantages in specific scenarios, with the following key insights:
\begin{itemize}
\item \emph{Pre-train a general policy, then post-train task-specific variants:}
  This widely adopted strategy has proven effective in domains such as language modeling~\cite{devlin2019bert} and robotics~\cite{black2024pi_0}.
  Fine-tuning enables policies to specialize for deployment environments, achieving performance not attainable with general models alone.
  We demonstrate that the same principle is crucial in MAPF, where the structural characteristics of each map strongly influence policy effectiveness.
\item \emph{Override policy outputs if necessary:}
  Neural MAPF policies are inherently imperfect and may lead to deadlocks or livelocks in practice.
  Fortunately, such pathological behaviors can be readily detected by examining recent state histories.
  A fallback is always available when detected, allowing for a return to standard MAPF methods without neural guidance.
  This corrective layer, largely overlooked in prior studies, proves essential for achieving high performance in hybrid approaches.
 \item \emph{Imitation that scales:}
    In our dense scenarios, while \lacamthree quickly produces near‑optimal trajectories for lower agent densities, it cannot reach the same quality for higher agent density instances within the \SI{30}{\second} budget.
    \magatstar is trained on the near‑optimal trajectories generated for the lower agent densities and is able to learn the rules which generalize to higher agent densities, even where the teacher itself falters.
    We hypothesize that, using the guidance from \magatstar, \lagat is able to visit higher‑quality states, leading to better solutions than \lacamthree~in~the~densest challenging setups.

\end{itemize}
Combined with other carefully engineered aspects, \lagat redefines the Pareto frontier that marks the speed–quality tradeoff established by leading search-based solvers.

\paragraph{When is LaGAT appealing?}
Current \lagat excels in scenarios that meet the following conditions:
\emph{(i)}~relatively small maps (e.g., $20$$\times$$20$),
\emph{(ii)}~obstacle-rich environments, and
\emph{(iii)}~high agent density.
For larger or less constrained maps, mere greedy collision-avoidance planners (like PIBT) along with global awareness~\cite{zhang2024guidance,kato2025congestion} are often sufficient to achieve near-optimal performance.
For maps much larger than the communication and observation ranges of decentralized policies, like \magatstar, the local nature of these policies can lead to suboptimal solutions, leaving room for future investigation in this area.
Moreover, agent density is a critical factor in determining instance difficulty: when the density is low, advanced MAPF solvers can already find solutions efficiently, leaving little room for neural components to offer practical benefits given their additional inference overhead.

\paragraph{Solidifying Neuro-Guided Multi-Agent Search.}
Our study gives positive evidence that combining search with neural guidance can achieve frontline performance in MAPF.
\lagat has two more desirable properties: \emph{(i)}~unlike purely learned approaches, it offers completeness guarantees, and \emph{(ii)}~the neural heuristic can be swapped in plug-and-play fashion, depending on the application. For these reasons, we believe this neuro-guided approach, where decentralized neural policies guide search, is a promising and flexible foundation for addressing broader classes of multi-agent coordination problems beyond MAPF.

\section*{Acknowledgments}
This research was funded in part by European Research Council (ERC)
Project 949940 (gAIa), JST ACT-X (JPMJAX22A1), and JST PRESTO (JPMJPR2513).

\bibliography{sty/ref-macro,main}

\begin{thebibliography}{40}
\providecommand{\natexlab}[1]{#1}

\bibitem[{Alam et~al.(2025)Alam, Mahmud, Mamun-Or-Rashid, and
  Khan}]{alam2025optimizing}
Alam, M.~A.; Mahmud, S.; Mamun-Or-Rashid, M.; and Khan, M.~M. 2025.
\newblock Optimizing Node Selection in Search Based Multi-Agent Path Finding.
\newblock \emph{Autonomous Agents and Multi-Agent Systems (JAAMAS)}.

\bibitem[{Alkazzi and Okumura(2024)}]{alkazzi2024comprehensive}
Alkazzi, J.-M.; and Okumura, K. 2024.
\newblock A comprehensive review on leveraging machine learning for multi-agent
  path finding.
\newblock \emph{IEEE Access}.

\bibitem[{Andreychuk et~al.(2025)Andreychuk, Yakovlev, Panov, and
  Skrynnik}]{andreychuk2025mapf}
Andreychuk, A.; Yakovlev, K.; Panov, A.; and Skrynnik, A. 2025.
\newblock MAPF-GPT: Imitation learning for multi-agent pathfinding at scale.
\newblock In \emph{Proceedings of AAAI Conference on Artificial Intelligence
  (AAAI)}.

\bibitem[{Barer et~al.(2014)Barer, Sharon, Stern, and Felner}]{Barer-ECBS}
Barer, M.; Sharon, G.; Stern, R.; and Felner, A. 2014.
\newblock Suboptimal variants of the conflict-based search algorithm for the
  multi-agent pathfinding problem.
\newblock In \emph{Proceedings of Annual Symposium on Combinatorial Search
  (SoCS)}.

\bibitem[{Black et~al.(2024)Black, Brown, Driess, Esmail, Equi, Finn, Fusai,
  Groom, Hausman, Ichter et~al.}]{black2024pi_0}
Black, K.; Brown, N.; Driess, D.; Esmail, A.; Equi, M.; Finn, C.; Fusai, N.;
  Groom, L.; Hausman, K.; Ichter, B.; et~al. 2024.
\newblock $\pi_0$: A Vision-Language-Action Flow Model for General Robot
  Control.
\newblock \emph{arXiv preprint}.

\bibitem[{Chen et~al.(2024)Chen, Harabor, Li, and Stuckey}]{chen2024traffic}
Chen, Z.; Harabor, D.; Li, J.; and Stuckey, P.~J. 2024.
\newblock Traffic flow optimisation for lifelong multi-agent path finding.
\newblock In \emph{Proceedings of AAAI Conference on Artificial Intelligence
  (AAAI)}.

\bibitem[{Devlin et~al.(2019)Devlin, Chang, Lee, and
  Toutanova}]{devlin2019bert}
Devlin, J.; Chang, M.-W.; Lee, K.; and Toutanova, K. 2019.
\newblock Bert: Pre-training of deep bidirectional transformers for language
  understanding.
\newblock In \emph{Proc. NAACL-HLT}.

\bibitem[{Jiang et~al.(2025)Jiang, Wang, Veerapaneni, Duhan, Sartoretti, and
  Li}]{jiang2025deploying}
Jiang, H.; Wang, Y.; Veerapaneni, R.; Duhan, T.; Sartoretti, G.; and Li, J.
  2025.
\newblock Deploying Ten Thousand Robots: Scalable Imitation Learning for
  Lifelong Multi-Agent Path Finding.
\newblock In \emph{Proceedings of IEEE International Conference on Robotics and
  Automation (ICRA)}.

\bibitem[{Kato et~al.(2025)Kato, Okumura, Sasaki, and
  Yokomachi}]{kato2025congestion}
Kato, T.; Okumura, K.; Sasaki, Y.; and Yokomachi, N. 2025.
\newblock Congestion Mitigation Path Planning for Large-Scale Multi-Agent
  Navigation in Dense Environments.
\newblock \emph{IEEE Robotics and Automation Letters (RA-L)}.

\bibitem[{Li et~al.(2021{\natexlab{a}})Li, Chen, Harabor, Stuckey, and
  Koenig}]{li2021anytime}
Li, J.; Chen, Z.; Harabor, D.; Stuckey, P.; and Koenig, S. 2021{\natexlab{a}}.
\newblock Anytime multi-agent path finding via large neighborhood search.
\newblock In \emph{Proceedings of International Joint Conference on Artificial
  Intelligence (IJCAI)}.

\bibitem[{Li et~al.(2022)Li, Chen, Harabor, Stuckey, and Koenig}]{li2022mapf}
Li, J.; Chen, Z.; Harabor, D.; Stuckey, P.~J.; and Koenig, S. 2022.
\newblock MAPF-LNS2: Fast repairing for multi-agent path finding via large
  neighborhood search.
\newblock In \emph{Proceedings of AAAI Conference on Artificial Intelligence
  (AAAI)}.

\bibitem[{Li et~al.(2020)Li, Gama, Ribeiro, and Prorok}]{li2020graph}
Li, Q.; Gama, F.; Ribeiro, A.; and Prorok, A. 2020.
\newblock Graph Neural Networks for Decentralized Multi-Robot Path Planning.
\newblock In \emph{Proceedings of IEEE/RSJ International Conference on
  Intelligent Robots and Systems (IROS)}.

\bibitem[{Li et~al.(2021{\natexlab{b}})Li, Lin, Liu, and
  Prorok}]{li2021message}
Li, Q.; Lin, W.; Liu, Z.; and Prorok, A. 2021{\natexlab{b}}.
\newblock Message-aware graph attention networks for large-scale multi-robot
  path planning.
\newblock \emph{IEEE Robotics and Automation Letters (RA-L)}.

\bibitem[{Loshchilov and Hutter(2019)}]{Loshchilov-AdamW}
Loshchilov, I.; and Hutter, F. 2019.
\newblock Decoupled weight decay regularization.
\newblock In \emph{Proceedings of the International Conference on Learning and
  Representation (ICLR)}.

\bibitem[{Ma, Luo, and Pan(2021)}]{Ma-DCC}
Ma, Z.; Luo, Y.; and Pan, J. 2021.
\newblock Learning selective communication for multi-agent path finding.
\newblock \emph{IEEE Robotics and Automation Letters (RA-L)}.

\bibitem[{Okumura(2023{\natexlab{a}})}]{okumura2023lacam2}
Okumura, K. 2023{\natexlab{a}}.
\newblock Improving LaCAM for Scalable Eventually Optimal Multi-Agent
  Pathfinding.
\newblock In \emph{Proceedings of International Joint Conference on Artificial
  Intelligence (IJCAI)}.

\bibitem[{Okumura(2023{\natexlab{b}})}]{okumura2023lacam}
Okumura, K. 2023{\natexlab{b}}.
\newblock LaCAM: Search-Based Algorithm for Quick Multi-Agent Pathfinding.
\newblock In \emph{Proceedings of AAAI Conference on Artificial Intelligence
  (AAAI)}.

\bibitem[{Okumura(2024)}]{okumura2024lacam3}
Okumura, K. 2024.
\newblock Engineering LaCAM$^\ast$: Towards Real-Time, Large-Scale, and
  Near-Optimal Multi-Agent Pathfinding.
\newblock In \emph{Proceedings of International Joint Conference on Autonomous
  Agents \& Multiagent Systems (AAMAS)}.

\bibitem[{Okumura et~al.(2022)Okumura, Machida, Défago, and
  Tamura}]{okumura2022priority}
Okumura, K.; Machida, M.; Défago, X.; and Tamura, Y. 2022.
\newblock Priority Inheritance with Backtracking for Iterative Multi-agent Path
  Finding.
\newblock \emph{Artificial Intelligence (AIJ)}.

\bibitem[{Okumura and Nagai(2025)}]{okumura2025lightweight}
Okumura, K.; and Nagai, H. 2025.
\newblock Lightweight and Effective Preference Construction in PIBT for
  Large-Scale Multi-Agent Pathfinding.
\newblock In \emph{Proceedings of Annual Symposium on Combinatorial Search
  (SoCS)}.

\bibitem[{Okumura, Tamura, and D\'{e}fago(2021)}]{okumura2021iterative}
Okumura, K.; Tamura, Y.; and D\'{e}fago, X. 2021.
\newblock Iterative Refinement for Real-Time Multi-Robot Path Planning.
\newblock In \emph{Proceedings of IEEE/RSJ International Conference on
  Intelligent Robots and Systems (IROS)}.

\bibitem[{Prorok et~al.(2021)Prorok, Blumenkamp, Li, Kortvelesy, Liu, and
  Stump}]{prorok2021holy}
Prorok, A.; Blumenkamp, J.; Li, Q.; Kortvelesy, R.; Liu, Z.; and Stump, E.
  2021.
\newblock The holy grail of multi-robot planning: Learning to generate
  online-scalable solutions from offline-optimal experts.
\newblock In \emph{Proceedings of International Joint Conference on Autonomous
  Agents \& Multiagent Systems (AAMAS)}.

\bibitem[{Ross, Gordon, and Bagnell(2011)}]{ross2011reduction}
Ross, S.; Gordon, G.; and Bagnell, D. 2011.
\newblock A reduction of imitation learning and structured prediction to
  no-regret online learning.
\newblock In \emph{Proceedings of the International Conference on Artificial
  Intelligence and Statistics (AISTATS)}.

\bibitem[{Sartoretti et~al.(2019)Sartoretti, Kerr, Shi, Wagner, Kumar, Koenig,
  and Choset}]{sartoretti2019primal}
Sartoretti, G.; Kerr, J.; Shi, Y.; Wagner, G.; Kumar, T.~S.; Koenig, S.; and
  Choset, H. 2019.
\newblock Primal: Pathfinding via reinforcement and imitation multi-agent
  learning.
\newblock \emph{IEEE Robotics and Automation Letters (RA-L)}.

\bibitem[{Shankar, Okumura, and Prorok(2025)}]{shankar2025lf}
Shankar, A.; Okumura, K.; and Prorok, A. 2025.
\newblock LF: Online Multi-Robot Path Planning Meets Optimal Trajectory
  Control.
\newblock \emph{arXiv preprint arXiv:2507.11464}.

\bibitem[{Shen et~al.(2023)Shen, Chen, Cheema, Harabor, and
  Stuckey}]{shen2023tracking}
Shen, B.; Chen, Z.; Cheema, M.~A.; Harabor, D.~D.; and Stuckey, P.~J. 2023.
\newblock Tracking progress in multi-agent path finding.
\newblock In \emph{Proceedings of International Conference on Automated
  Planning and Scheduling (ICAPS)}.

\bibitem[{Silver et~al.(2016)Silver, Huang, Maddison, Guez, Sifre, Van
  Den~Driessche, Schrittwieser, Antonoglou, Panneershelvam, Lanctot
  et~al.}]{silver2016mastering}
Silver, D.; Huang, A.; Maddison, C.~J.; Guez, A.; Sifre, L.; Van Den~Driessche,
  G.; Schrittwieser, J.; Antonoglou, I.; Panneershelvam, V.; Lanctot, M.;
  et~al. 2016.
\newblock Mastering the game of Go with deep neural networks and tree search.
\newblock \emph{Nature}.

\bibitem[{Silver et~al.(2017)Silver, Schrittwieser, Simonyan, Antonoglou,
  Huang, Guez, Hubert, Baker, Lai, Bolton et~al.}]{silver2017mastering}
Silver, D.; Schrittwieser, J.; Simonyan, K.; Antonoglou, I.; Huang, A.; Guez,
  A.; Hubert, T.; Baker, L.; Lai, M.; Bolton, A.; et~al. 2017.
\newblock Mastering the game of go without human knowledge.
\newblock \emph{Nature}.

\bibitem[{Skrynnik et~al.(2025)Skrynnik, Andreychuk, Borzilov, Chernyavskiy,
  Yakovlev, and Panov}]{skrynnik2025pogema}
Skrynnik, A.; Andreychuk, A.; Borzilov, A.; Chernyavskiy, A.; Yakovlev, K.; and
  Panov, A. 2025.
\newblock Pogema: A benchmark platform for cooperative multi-agent pathfinding.
\newblock In \emph{Proceedings of the International Conference on Learning and
  Representation (ICLR)}.

\bibitem[{Stern et~al.(2019)Stern, Sturtevant, Felner, Koenig, Ma, Walker, Li,
  Atzmon, Cohen, Kumar et~al.}]{stern2019def}
Stern, R.; Sturtevant, N.; Felner, A.; Koenig, S.; Ma, H.; Walker, T.; Li, J.;
  Atzmon, D.; Cohen, L.; Kumar, T.; et~al. 2019.
\newblock Multi-Agent Pathfinding: Definitions, Variants, and Benchmarks.
\newblock In \emph{Proceedings of Annual Symposium on Combinatorial Search
  (SoCS)}.

\bibitem[{Tan et~al.(2025)Tan, Luo, Li, and Ma}]{tan2025reevaluation}
Tan, J.; Luo, Y.; Li, J.; and Ma, H. 2025.
\newblock Reevaluation of Large Neighborhood Search for MAPF: Findings and
  Opportunities.
\newblock In \emph{Proceedings of Annual Symposium on Combinatorial Search
  (SoCS)}.

\bibitem[{Veerapaneni et~al.(2025)Veerapaneni, Jakobsson, Ren, Kim, Li, and
  Likhachev}]{veerapaneni2025work}
Veerapaneni, R.; Jakobsson, A.; Ren, K.; Kim, S.; Li, J.; and Likhachev, M.
  2025.
\newblock Work Smarter Not Harder: Simple Imitation Learning with CS-PIBT
  Outperforms Large Scale Imitation Learning for MAPF.
\newblock In \emph{Proceedings of IEEE International Conference on Robotics and
  Automation (ICRA)}.

\bibitem[{Veerapaneni et~al.(2024)Veerapaneni, Wang, Ren, Jakobsson, Li, and
  Likhachev}]{veerapaneni2024improving}
Veerapaneni, R.; Wang, Q.; Ren, K.; Jakobsson, A.; Li, J.; and Likhachev, M.
  2024.
\newblock Improving Learnt Local MAPF Policies with Heuristic Search.
\newblock In \emph{Proceedings of International Conference on Automated
  Planning and Scheduling (ICAPS)}.

\bibitem[{Wang et~al.(2025)Wang, Duhan, Li, and Sartoretti}]{wang2025lns2+}
Wang, Y.; Duhan, T.; Li, J.; and Sartoretti, G. 2025.
\newblock LNS2+ RL: Combining multi-agent reinforcement learning with large
  neighborhood search in multi-agent path finding.
\newblock In \emph{Proceedings of AAAI Conference on Artificial Intelligence
  (AAAI)}.

\bibitem[{Wang et~al.(2023)Wang, Xiang, Huang, and Sartoretti}]{Wang-SCRIMP}
Wang, Y.; Xiang, B.; Huang, S.; and Sartoretti, G. 2023.
\newblock Scrimp: Scalable communication for reinforcement-and
  imitation-learning-based multi-agent pathfinding.
\newblock In \emph{Proceedings of IEEE/RSJ International Conference on
  Intelligent Robots and Systems (IROS)}.

\bibitem[{Yan and Wu(2024)}]{yan2024neural}
Yan, Z.; and Wu, C. 2024.
\newblock Neural neighborhood search for multi-agent path finding.
\newblock In \emph{Proceedings of the International Conference on Learning and
  Representation (ICLR)}.

\bibitem[{Yu et~al.(2023)Yu, Li, Gao, and Prorok}]{yu2023accelerating}
Yu, C.; Li, Q.; Gao, S.; and Prorok, A. 2023.
\newblock Accelerating multi-agent planning using graph transformers with
  bounded suboptimality.
\newblock In \emph{Proceedings of IEEE International Conference on Robotics and
  Automation (ICRA)}.

\bibitem[{Yukhnevich and Andreychuk(2025)}]{yukhnevich2025enhancing}
Yukhnevich, E.; and Andreychuk, A. 2025.
\newblock Enhancing PIBT via multi-action operations.
\newblock In \emph{League of Robot Runners Expo}.

\bibitem[{Zhang et~al.(2024{\natexlab{a}})Zhang, Chen, Harabor, Le~Bodic, and
  Stuckey}]{zhang2024planning}
Zhang, Y.; Chen, Z.; Harabor, D.; Le~Bodic, P.; and Stuckey, P.~J.
  2024{\natexlab{a}}.
\newblock Planning and execution in multi-agent path finding: Models and
  algorithms.
\newblock In \emph{Proceedings of International Conference on Automated
  Planning and Scheduling (ICAPS)}.

\bibitem[{Zhang et~al.(2024{\natexlab{b}})Zhang, Jiang, Bhatt, Nikolaidis, and
  Li}]{zhang2024guidance}
Zhang, Y.; Jiang, H.; Bhatt, V.; Nikolaidis, S.; and Li, J. 2024{\natexlab{b}}.
\newblock Guidance graph optimization for lifelong multi-agent path finding.
\newblock In \emph{Proceedings of International Joint Conference on Artificial
  Intelligence (IJCAI)}.

\end{thebibliography}
\appendix
\onecolumn
\setcounter{secnumdepth}{2}
\section*{Appendix}

\section{\leftline{Training Details of \magatstar}}
\magatstar is trained using a pipeline similar to that of \citet{li2021message}.
Every four epochs, we sample $500$ training instances from the original dataset, and solve them using the current \magatstar model.
For instances where the model fails, we invoke an online expert planner (\lacamthree) to generate ground-truth trajectories from the final configuration; these are then added to the training set.
The model is trained for $200$ epochs using the AdamW optimizer, with a batch size of $64$ and a learning rate that decays from $1\mathrm{e}{-3}$ to $1\mathrm{e}{-6}$ following a cosine annealing schedule.

For finetuning the model, we follow the same setup as above, but for $52$ epochs only.

\section{\leftline{Proof of \cref{thrm:dd}}}
\setcounter{theorem}{0}
\begin{theorem}
  LaCAM with \cref{algo:dd} is complete.
\end{theorem}
\begin{proofsketch}
According to~\cite{okumura2023lacam}, LaCAM is complete regardless of the preferences in PIBT.
\Cref{algo:dd} does not violate the structural assumptions of LaCAM, except for~\cref{algo:dd:reset}, which resets constraints.
For each search node, once an agent $i \in A$ is added to the \unguided set, it remains \unguided for the rest of the search.
Since $A$ is finite, the \unguided set cannot grow infinitely, and consequently, constraints cannot be reset infinitely often.
This ensures that, from any given configuration, all successor configurations will eventually be explored during the search, thereby establishing completeness.
\end{proofsketch}

\section{\leftline{Evaluation Setup for \cref{sec:result-policy}}}
\Cref{fig:benchmarks} shows the breakdown of tested $3,456$ instances for evaluating various neural policies in \cref{fig:success-by-runtime}.
{
  \newcommand{\entry}[6]{
  \begin{scope}[xshift=#1*0.15\linewidth]
    \node[] at (0, 1.7) {\scriptsize\mapname{#2}};
    \node[] at (0, 1.45) {\tiny #4$\times$#5};
    \node[] at (0, 0)
         {\includegraphics[width=0.15\linewidth]{fig/raw/neural_benchmark_maps/#3}};
    \node[] at (0, -1.6) {\scriptsize $[#6]$};
  \end{scope}
}
\begin{figure}[th!]
  \centering
  \begin{tikzpicture}
    \entry{0}{Sparse Warehouse}{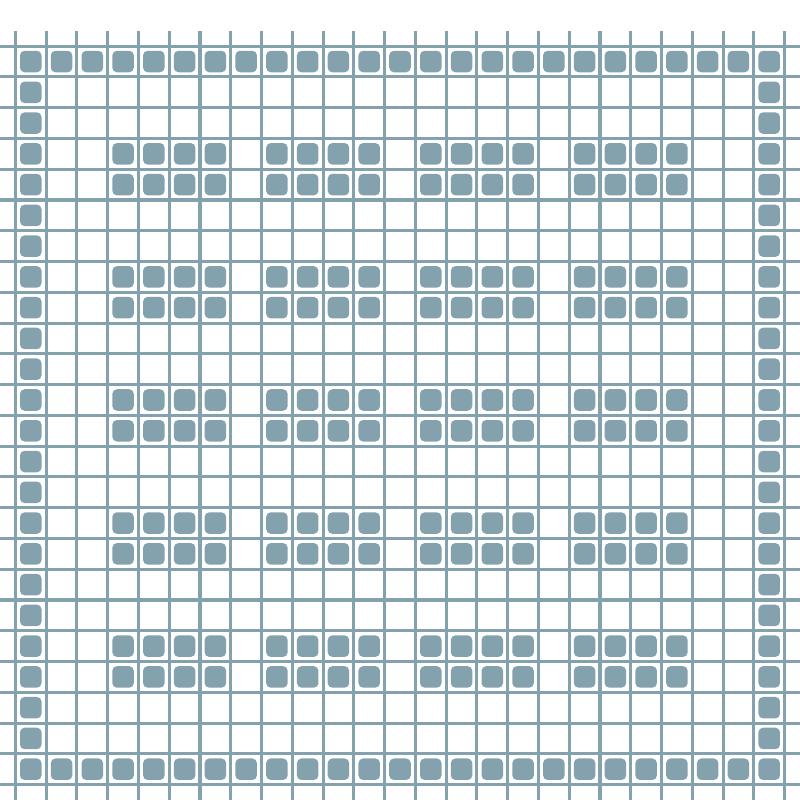}{23}{22}{32, 64, 96, 128, 160}
    \entry{1}{Dense Warehouse}{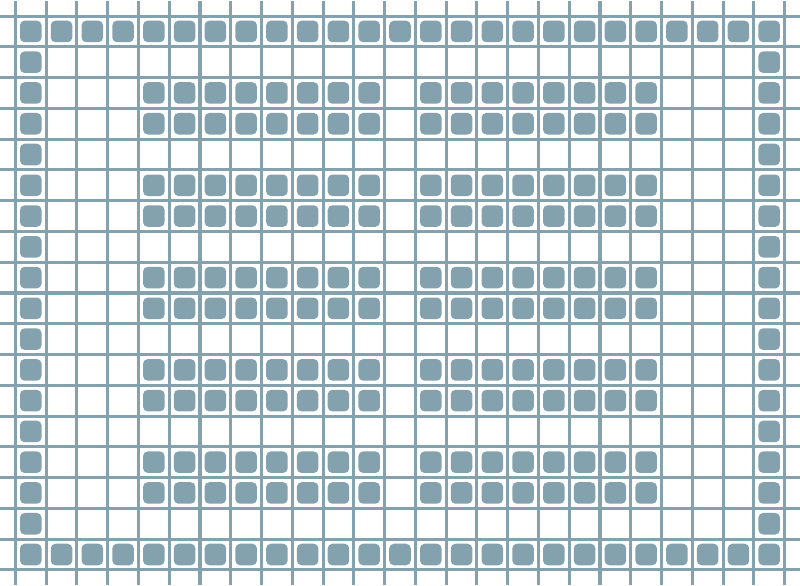}{23}{16}{32, 64, 96, 128}
    \entry{2}{Empty Room}{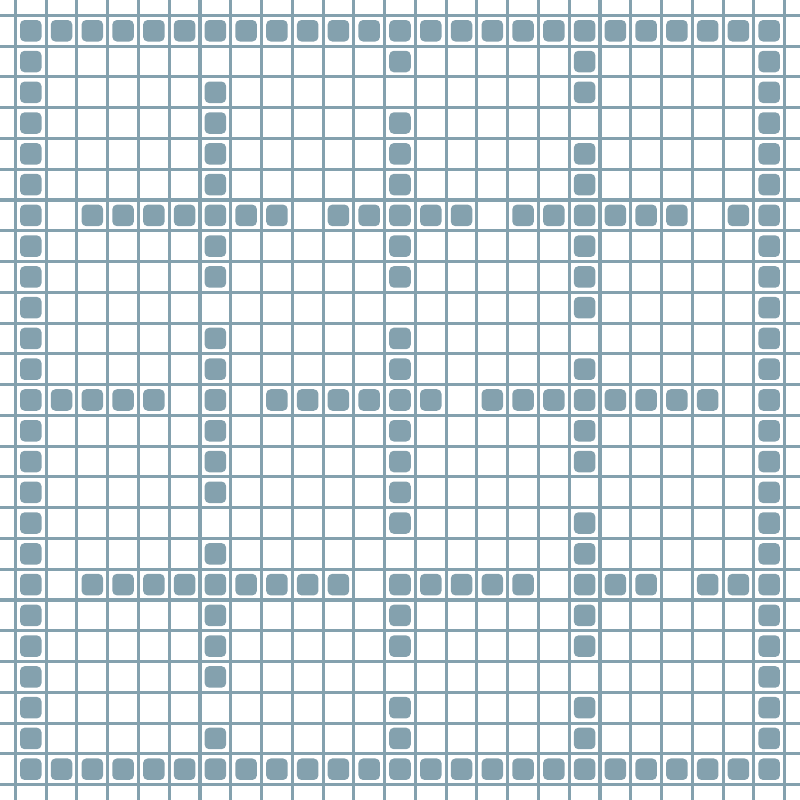}{20}{20}{32, 64, 96, 128, 160}
    \entry{3}{Dense Room}{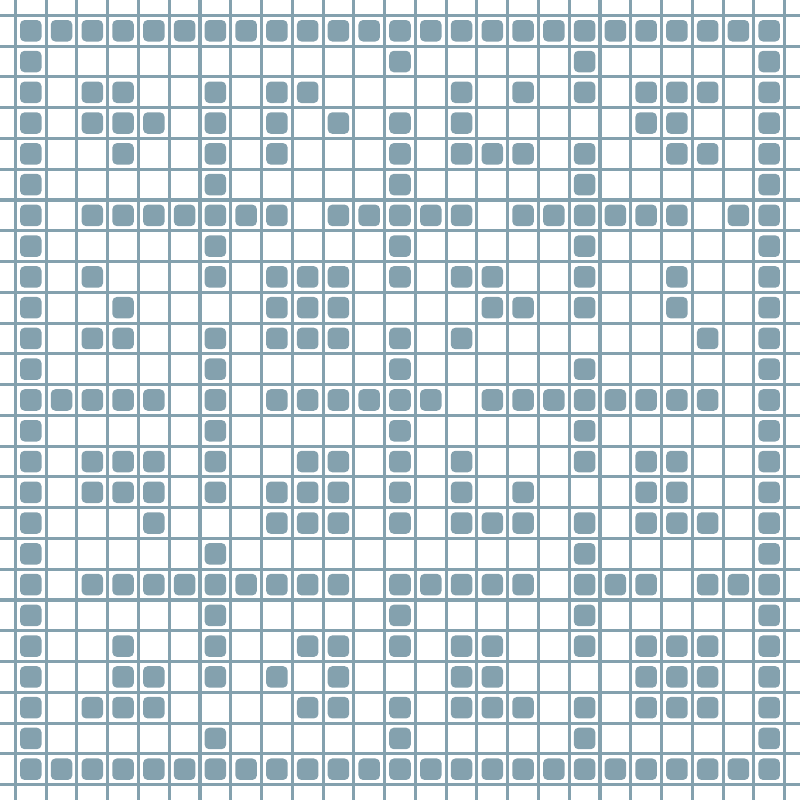}{23}{23}{32, 64, 96, 128}
    \entry{4}{Sparse Maze}{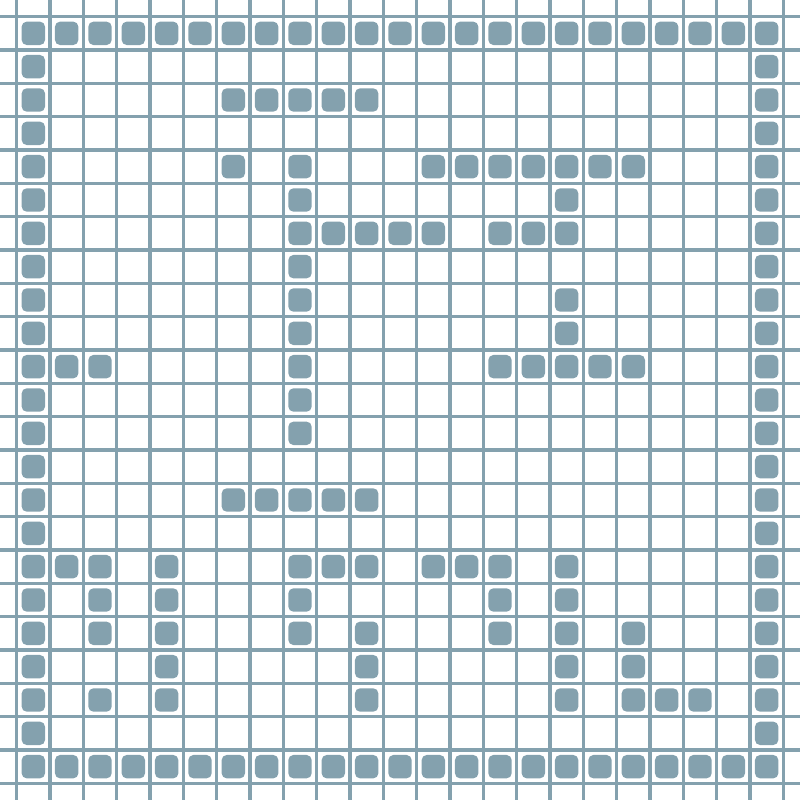}{23}{23}{32, 64, 96, 128, 160}
    \entry{5}{Dense Maze}{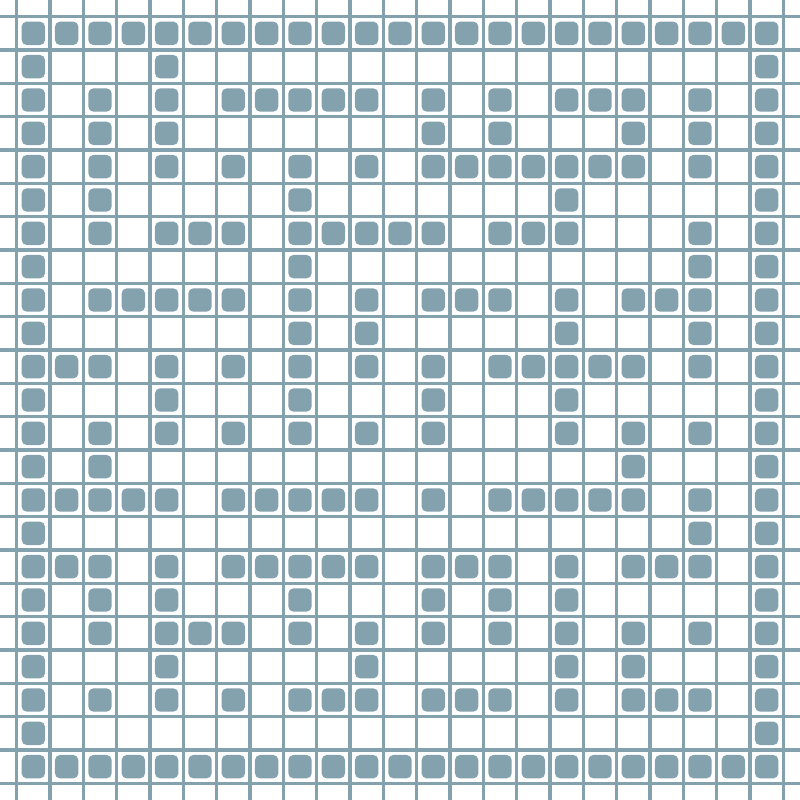}{21}{21}{32, 64, 96, 128}
    \node[anchor=east] at (-1.3, -1.6) {\scriptsize \#agents:};
  \end{tikzpicture}
  \caption{Breakdown of benchmarks instances used in \cref{fig:success-by-runtime}.}
  \label{fig:benchmarks}
\end{figure}
}

\section{\leftline{Extended Results of \cref{sec:result-main}}}

\Cref{fig:tradeoff2} compares real-time performance between \lagat and \lacamthree in the dense setups.
\Cref{fig:success_rate} presents planning success rate of each solver omitted from \cref{fig:main-eval}.
\Cref{fig:main-eval-supp} provides further experiments of \lagat on additional maps (32 instances for each setting).
As MAPF-GPT lacks real-time planning ability and is significantly slower than the others (c.f., see the main paper),
we have omitted the evaluation of MAPF-GPT.

{
\newcommand{\axSizeX}{5.0}
\newcommand{\axSizeY}{2.5}
\newcommand{\maxX}{33}
\newcommand{\maxY}{13.5}
\newcommand{\p}[2]{#1 / \maxX * \axSizeX, #2 / \maxY * \axSizeY}  
\begin{figure}[th!]
\centering
\scalebox{1.0}{
\begin{tikzpicture}
  \scriptsize
  %
  {
    \newcommand{\tick}[1]{
      node[pos=#1 / \maxX,below=1.0]{\scriptsize #1}
      node[pos=#1 / \maxX]{\tiny $|$}
    }
    \draw[black,->,line width=1pt] (0,0) -- (\axSizeX, 0)
    \tick{5}\tick{10}\tick{15}\tick{20}\tick{25}\tick{30}
    node[pos=0.0,below]{\scriptsize 0}
    node[pos=0.5,rotate=0,below=8.0]{runtime (\SI{}{\second})};
  };
  %
  {
    \newcommand{\tick}[1]{
      node[pos=#1 / \maxY,left]{\scriptsize #1}
      node[pos=#1 / \maxY]{\tiny $-$}
    }
    \draw[black,->,line width=1pt] (0,0) -- (0, \axSizeY)
    \tick{1}\tick{5}\tick{10}
    node[pos=0.5,rotate=90,above=16.0]{solution cost}
    node[pos=0.5,rotate=90,above=8.0]{\tiny(sum-of-costs/LB)};
  }
  %
  {
    \coordinate (c-lacam3-init) at (\p{4.43}{13.08});
    \coordinate (c-lacam3-10) at (\p{10}{9.61});
    \coordinate (c-lacam3-15) at (\p{15}{9.02});
    \coordinate (c-lacam3-20) at (\p{20}{8.26});
    \coordinate (c-lacam3-25) at (\p{25}{7.57});
    \coordinate (c-lacam3-30) at (\p{30}{7.33});
    \foreach \x in {c-lacam3-init,c-lacam3-10,c-lacam3-15,c-lacam3-20,c-lacam3-25,c-lacam3-30}
    \node[] at (\x) {$\bullet$};
    \draw[black]
    (c-lacam3-init) --
    (c-lacam3-10) --
    (c-lacam3-15) --
    (c-lacam3-20) --
    (c-lacam3-25) --
    (c-lacam3-30)
    node[right=1.0]{\lacamthree};
    \draw[lightgray,densely dashed] (c-lacam3-30) -- ++(-30 / \maxX * \axSizeX, 0);
  }
  {
    \coordinate (c-lagat-init) at (\p{9.92}{4.99});
    \coordinate (c-lagat-30) at (\p{30}{4.38});
    \node[red] at (c-lagat-init) {$\bigstar$};
    \node[red] at (c-lagat-30) {$\bigstar$};
    \node[red,below=1.0] at (c-lagat-init) {\lagat{\tiny -init}};
    \node[red,right=1.0] at (c-lagat-30) {\lagat{\tiny~(\SI{30}{\second})}};
    \draw[red](c-lagat-init) -- (c-lagat-30);
  }
  {
    \coordinate (c-magat) at (\p{3.06}{11.64});
    \node[] at (c-magat) {\tiny $\blacksquare$};
    \node[] at ($(c-magat) + (0.2, -0.25)$) {\magatstar};
  }
\end{tikzpicture}
}
\caption{
  Real-time performance comparison in the densest instances from \mapname{Dense Warehouse/Maze/Room}.
  The plot is based on average scores.
}
\label{fig:tradeoff2}
\end{figure}
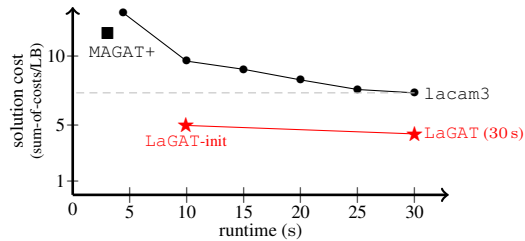
}

{
\newcommand{\entry}[7]{
  \begin{scope}[xshift=#1*0.16\linewidth]
  \node[anchor=north west] at (0, 0)
       {\includegraphics[width=0.16\linewidth]{fig/raw/main_eval/#3_success}};
  \node[anchor=center] at (1.65, -0.05) {\scriptsize\mapname{#2}};
  \end{scope}
}
\begin{figure*}[th!]
  \centering
  \begin{tikzpicture}
    \entry{0}{Dense Warehouse}{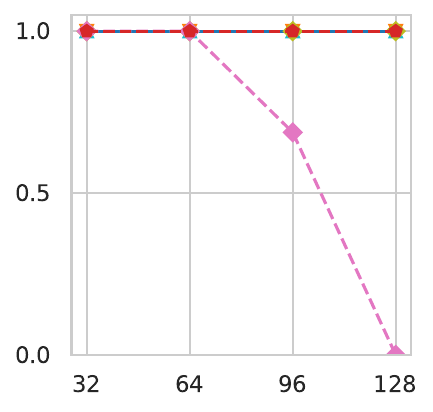}{23}{16}{0}{0.08}
    \entry{1}{Dense Room}{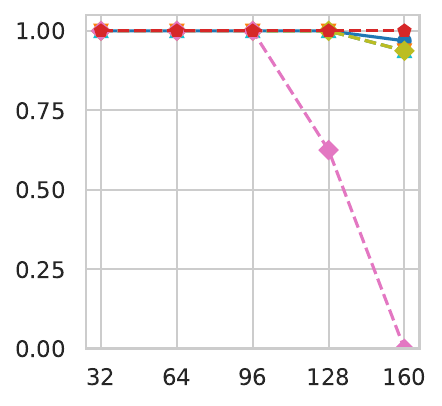}{23}{23}{-0.0}{0.06}
    \entry{2}{Dense Maze}{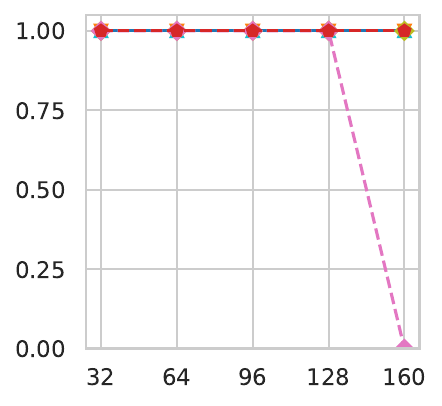}{21}{21}{-0.1}{0.06}
    \entry{3}{Empty Room}{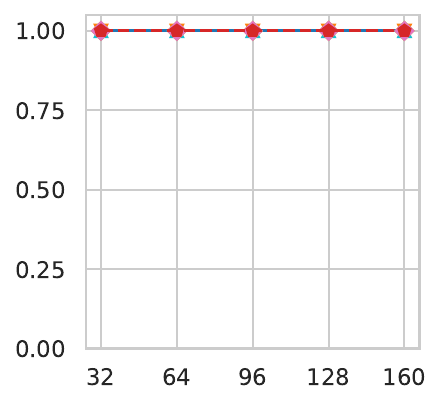}{23}{23}{0.05}{0.06}
    \entry{4}{ost003d}{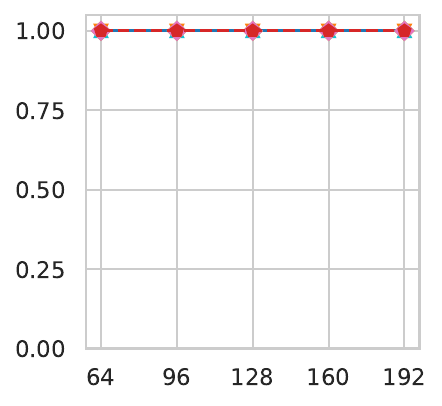}{194}{194}{0.05}{0.06}
    \entry{5}{Empty Map}{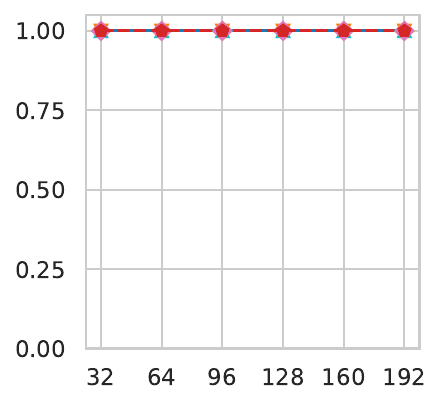}{20}{20}{0.1}{0.06}
    %
    \node[rotate=90] at (0, -1.5) {\small success rate};
    \node[] at (0.1\linewidth, -3) {\small \#agents};
    \node[anchor=east] at (0.96\linewidth, -3)
         {\includegraphics[width=0.75\linewidth,clip,trim={0.4cm 0.3cm 0.4cm 0.3cm}]{fig/raw/main_eval/legend}};
  \end{tikzpicture}
  \caption{
    Planning success rate of \cref{fig:main-eval}.
  }
  \label{fig:success_rate}
\end{figure*}
}

{
\newcommand{\fheight}{0.16\linewidth}
\newcommand{\fwidth}{0.16\linewidth}

\newcommand{\entry}[5]{
  \begin{scope}[xshift=#1*0.16\linewidth]
    \node[] at (1.65, 2.4) {\scriptsize\mapname{#2}};
    \node[] at (1.65, 2.2) {\tiny #4$\times$#5};
    \node[anchor=center,fill=white,inner sep=0pt] at (1.65, 1.0)
         {\includegraphics[width=0.12\linewidth]{fig/raw/appendix_eval/#3}};
    \node[anchor=north west] at (0, 0)
         {\includegraphics[width=\fwidth]{fig/raw/appendix_eval/#3_success}};
    \node[anchor=north west] at (0, -2.8)
         {\includegraphics[width=\fwidth,height=\fheight]{fig/raw/appendix_eval/#3_per_map_runtime}};
    \node[anchor=north west] at (0, -5.6)
         {\includegraphics[width=\fwidth,height=\fheight]{fig/raw/appendix_eval/#3_sum_of_costs}};
  \end{scope}
}

\begin{figure*}[th!]
  \centering
  \begin{tikzpicture}
    \entry{0}{lak303d}{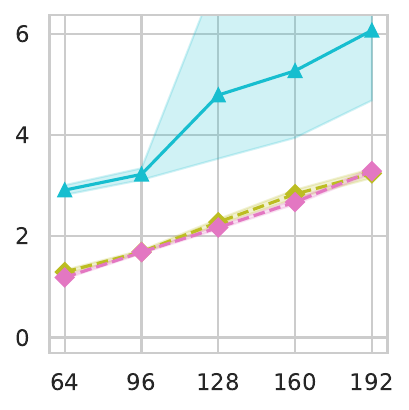}{194}{194}
    \entry{1}{Sortation}{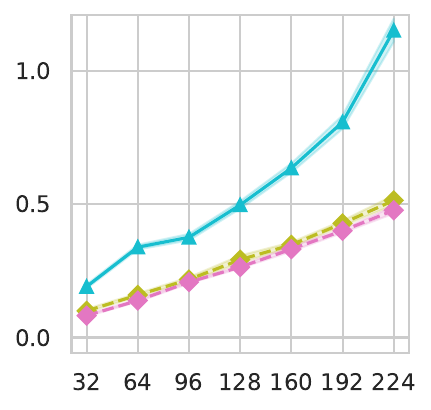}{33}{57}
    \entry{2}{Horizontal Stripes}{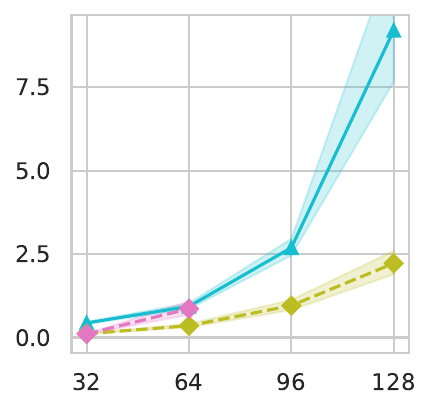}{23}{22}
    \entry{3}{Dense Random}{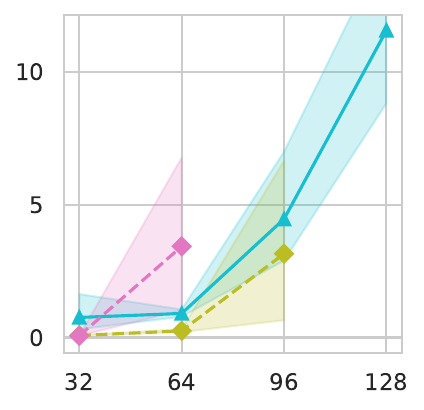}{20}{20}
    %
    \node[rotate=90] at (0, -1.5) {\small Success rate};
    \node[rotate=90] at (0, -4.3) {\small Runtime (\SI{}{\second})};
    \node[rotate=90] at (0, -7.0) {\small SoC / LB};
  \end{tikzpicture}
  \\
  {\small \#agents}
  \smallskip\\
  \includegraphics[width=0.75\linewidth,clip,trim={0.4cm 0.3cm 0.4cm 0.3cm}]{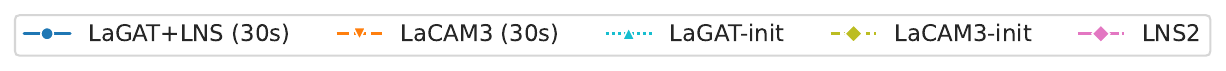}
  \caption{
    Evaluation on additional maps.
  }
  \label{fig:main-eval-supp}
\end{figure*}
}

In \cref{fig:main-eval-supp}, we observe similar trends as noted in \cref{sec:result-main}.
\mapname{lak303d} is a large, sparse map similar to \mapname{ost003d}, and we notice
the same identical pattern of both \lagat and \lacamthree getting near-optimal solutions,
with the initial solution of \lagat being slightly worse than \lacamthree's, but
the final solution being slightly better. In the \mapname{Sortation} map, we observe
similar trends as observed with the \mapname{Empty Map}, which is expected
as both are small (relative to \mapname{lak303d} and \mapname{ost003d}) but sparse maps.
For the \mapname{Horizontal Stripes} map, we observe that \lagat gets better quality
final and initial solutions than \lacamthree, in line with the results for the other
dense maps in \cref{sec:result-main}.

Notably, for the \mapname{Dense Random} map, we notice that for the highest agent
density situation, \lagat achieves $93.75\%$ success rate within \SI{30}{\second},
while \lacamthree achieves only $75\%$ success rate. Thus, \lagat is the only solver that
is able to achieve high success rate in this map within a reasonable runtime.
This means that \lagat not only is able to provide higher quality solutions
than \lacamthree, but it is also be solve scenarios where \lacamthree fails
to find a solution within a reasonable time limit.

\section{\leftline{Extended Results of \cref{sec:result-ablation}}}
\Cref{fig:ablation-bar-graph-combined} shows the results of the ablation study
on the densest scenarios for the first three maps from \cref{fig:main-eval}
combined.
This illustrates that to outperform \lacamthree with both a high success rate and a low solution cost, all components---pre-training, fine-tuning, neural guidance, deadlock detection and search---are necessary.
{
  \begin{figure}[tp!]
    \centering
    \includegraphics[width=0.7\linewidth]{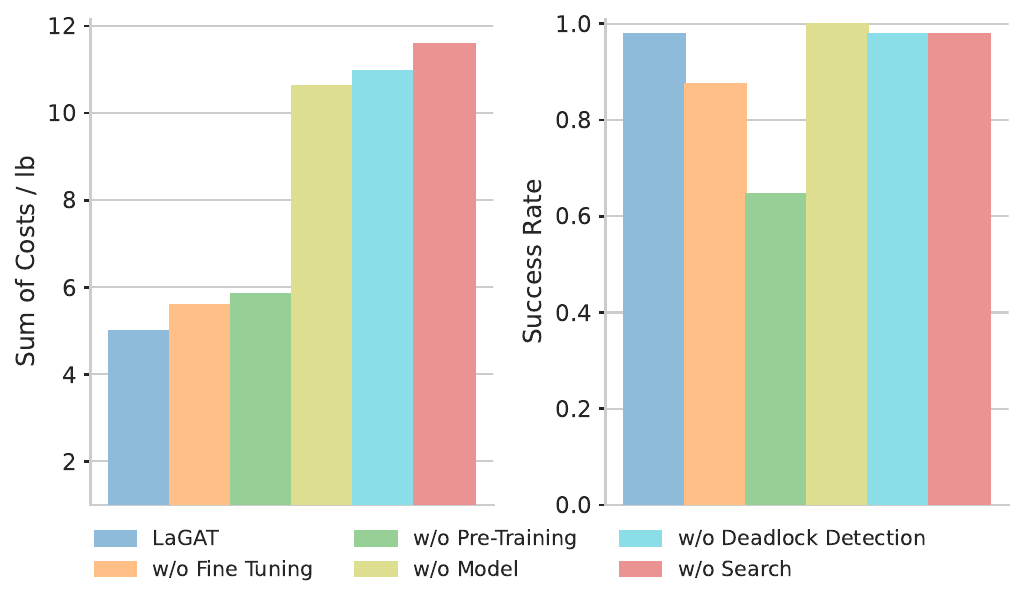}
    \caption{
      Ablation study on the densest scenarios in \cref{fig:main-eval}.
    }
    \label{fig:ablation-bar-graph-combined}
  \end{figure}
}

\end{document}